\documentclass[accepted]{uai2025} 
\usepackage[american]{babel}
\usepackage{natbib} 
\usepackage{mathtools} 
\usepackage{booktabs} 
\usepackage{tikz}

\usepackage{algorithm}
\usepackage{algorithmic}
\usepackage{subfigure}
\usepackage{multirow}
\usepackage{multicol}
\usepackage{amsmath}
\usepackage{amssymb}

\title{FlightPatchNet: Multi-Scale Patch Network with \\Differential Coding for Flight Trajectory Prediction}

\author[1]{Lan Wu}
\author[1]{Xuebin Wang  \thanks{coauthor}}
\author[1]{Ruijuan Chu}
\author[1]{Guangyi Liu}
\author[1]{Jing Zhang}
\author[1]{Linyu Wang \thanks{corresponding author}}

\affil[1]{%
    Information Engineering University\\
    Zhengzhou, Henan, China
}
  
\begin{document}
\maketitle

\begin{abstract}
  Accurate multi-step flight trajectory prediction plays an important role in Air Traffic Control, which can ensure the safety of air transportation. Two main issues limit the flight trajectory prediction performance of existing works. The first issue is the negative impact on prediction accuracy caused by the significant differences in data range. The second issue is that real-world flight trajectories involve underlying temporal dependencies, and most existing methods fail to reveal the hidden complex temporal variations and extract features from one single time scale. To address the above issues, we propose FlightPatchNet, a multi-scale patch network with differential coding for flight trajectory prediction. Specifically, FlightPatchNet first utilizes differential coding to encode the original values of longitude and latitude into first-order differences and generates embeddings for all variables at each time step. Then, global temporal attention is introduced to explore the dependencies between different time steps. To fully explore the diverse temporal patterns in flight trajectories, a multi-scale patch network is delicately designed to serve as the backbone. The multi-scale patch network exploits stacked patch mixer blocks to capture inter- and intra-patch dependencies under different time scales, and further integrates multi-scale temporal features across different scales and variables. Finally, FlightPatchNet ensembles multiple predictors to make direct multi-step prediction. Extensive experiments on ADS-B datasets demonstrate that our model outperforms the competitive baselines.
\end{abstract}

\section{Introduction}
Flight Trajectory Prediction (FTP) is an essential task in the Air Traffic Control (ATC) procedure, which can be applied to various scenarios such as air traffic flow prediction \citep{Abadi6878453, LIN2019105113}, aircraft conflict detection \citep{AdeepGaussian}, and arrival time estimation \citep{WANG2018280}. Accurate FTP can ensure the safety of air transportation and improve real-time airspace management \citep{LIN8846596, Shi9136843}. Generally, FTP tasks can be divided into three categories: long-term \citep{Jeong8190764, Runle7999617}, medium-term \citep{Yuan7554828, Chen2016ShortmediumtermPF}, and short-term \citep{huang2017short, duan2018unified}. Among them, short-term trajectory prediction has the greatest impact on ATC and is increasingly in demand for air transportation. In this paper, we mainly focus on the short-term FTP task, which aims to predict future flight trajectories based on historical observations.
 
In the ATC domain, multi-step trajectory prediction can provide more practical applications than single-step prediction \citep{LIN8846596}. It can be divided into Iterated Multi-Step (IMS) prediction and Direct Multi-Step (DMS) prediction. IMS-based methods \citep{Yan6972562, Zhang2023FlightTP, Guo2023FlightBERT} make multi-step prediction recursively, which learns a single-step model and iteratively applies the predicted values as observations to forecast the next trajectory point. Due to the error accumulation problem and the step-by-step prediction scheme, this type of methods usually fails in multi-step prediction and has poor real-time performance. By contrast, DMS-based methods \citep{wuhan2023bi, Guo2023FlightBERT++} can directly generate future trajectory points at once, which can tackle the error accumulation problem and improve prediction efficiency. Therefore, this paper performs the short-term FTP task in DMS way.  

However, two main issues are not well addressed in existing works \citep{Yan6972562, Zhang2023FlightTP, wuhan2023bi, Guo2023FlightBERT++},  limiting the trajectory prediction performance. The first issue is the negative impact on prediction accuracy caused by the significant differences in data range. In general, longitude and latitude are denoted by degree but altitude is by meter. Since one degree is approximately 111 kilometers, the data range of longitude and latitude are extremely different from that of altitude. Some previous works \citep{LSTM8489734, CNN-LSTM9145522} directly utilized normalization algorithms to scale variables into the same range, e.g., from 0 to 1. However, the actual prediction errors could be large for FTP tasks when evaluated in raw data range (as shown in Table 2). FlightBERT \citep{Guo2023FlightBERT} and FlightBERT++ \citep{Guo2023FlightBERT++} proposed binary encoding (BE) representation to convert variables from rounded decimal numbers to binary vectors, which regards the FTP task as multiple binary classification problem. Although BE representation can avoid the vulnerability caused by normalization algorithms, one serious limitation is introduced: a high bit misclassification in binary will lead to a large absolute error in decimal.
\begin{figure}[h]
\centering 
    \subfigure[original series]{  
\centering    
\includegraphics[width=0.95\linewidth]{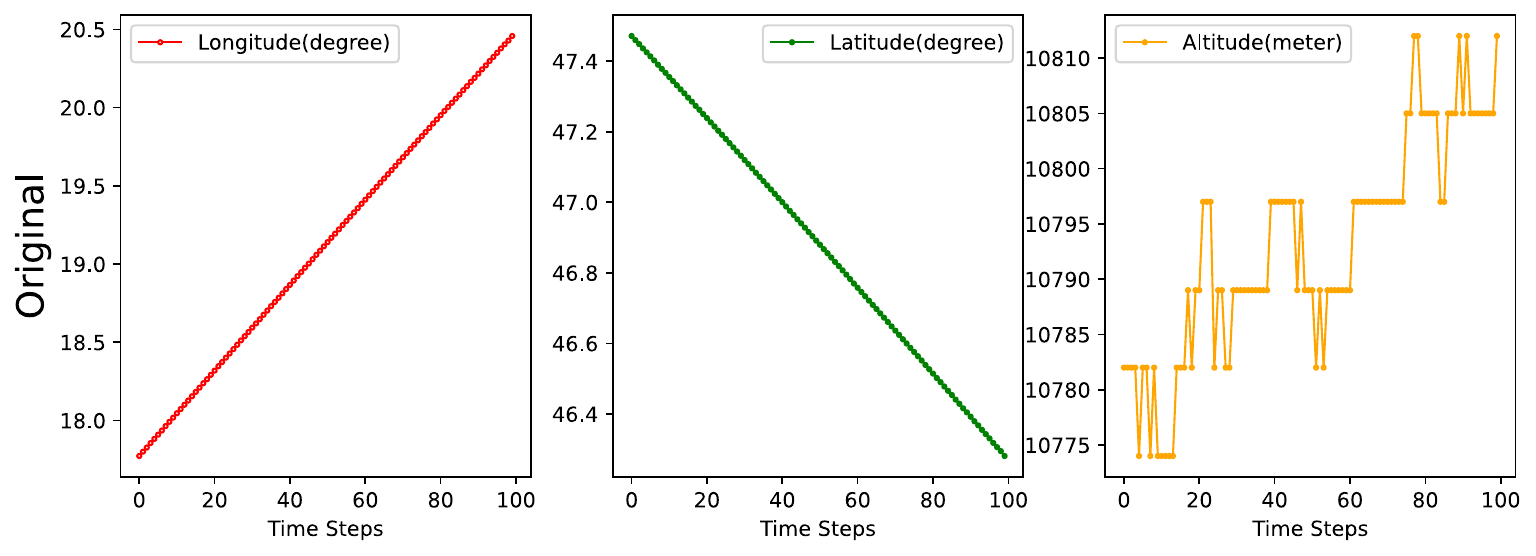}  
\label{fig:ori_traj}
}
    \subfigure[first-order difference series]{
\centering    
\includegraphics[width=0.95\linewidth]{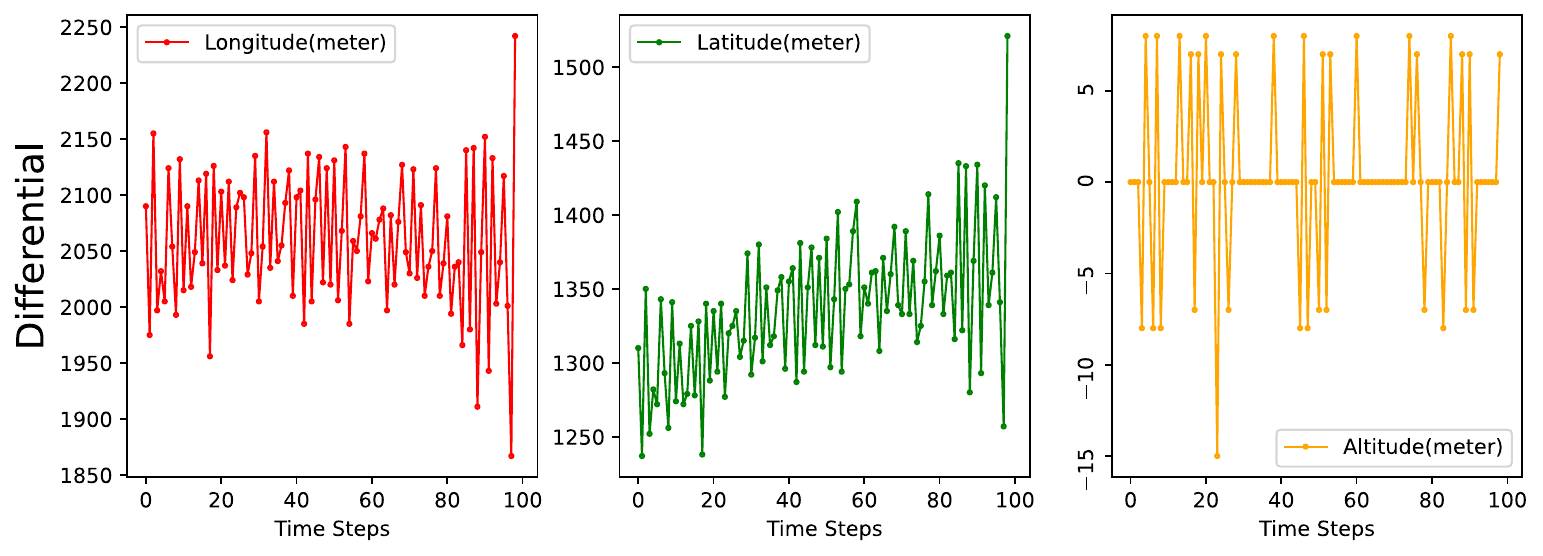}
\label{fig:diff_traj}
}
\captionsetup{font=small}
\caption{The original and first-order difference series in real-world flight trajectories.}   
\label{fig:traj}    
\end{figure}

The second issue is that real-world flight trajectories involve underlying temporal dependencies, and most existing methods \citep{Shi9136843, Guo2023FlightBERT, Guo2023FlightBERT++} fail to reveal the hidden complex temporal variations and  extract features from one single time scale. As shown in Figure~\ref{fig:traj}, the original series of longitude and latitude are over-smoothing and obscure abundant temporal variations, which can be observed from the first-order difference series. Besides, the temporal variation patterns of longitude and latitude are quite distinct from those of altitude which have an obvious global trend but suffer from intense local fluctuations. For example, slight turbulence can exert a significant influence on the altitude but produce a negligible effect on the longitude and latitude. A single-scale model cannot simultaneously capture both local temporal details and global trends \citep{wu2022timesnet, wang2023micn}. This calls for powerful multi-scale temporal modeling capacity. Furthermore, if the learned multi-scale temporal patterns are simply aggregated, the model is failed to focus on contributed patterns \citep{chen2023multi}. Meanwhile, it is essential to explore relationships across variables \citep{zhang2023crossformer, han2024capacity}, e.g., the velocity at current time step directly affects the location at next time step. Thus, scale-wise correlations and inter-variable relationships should be fully considered when modeling the multi-scale temporal patterns. 

Based on above analysis, this paper proposes a multi-scale patch network with differential coding (FlightPatchNet) to address above issues. Specifically, we utilize differential coding to encode the original values of longitude and latitude into first-order differences and retain the original values of other variables as inputs. Due to the dependencies between nearby and distant time steps, we introduce global temporal embedding to explore the correlations between time steps. Then, a multi-scale patch network is proposed to enable the ability of powerful and complete temporal modeling. The multi-scale patch network divides the trajectory series into patches of different sizes,  and exploits stacked patch mixer blocks to capture global trends across patches and local details within patches. To further promote the multi-scale temporal modeling capacity, a multi-scale aggregator is introduced to capture scale-wise correlations and inter-variable relationships. Finally, FlightPatchNet ensembles multiple predictors to make direct multi-step forecasting, which can benefit from complementary multi-scale temporal features and improve the generalization ability. The main contributions are summarized as follows: 
\begin{itemize}
    \item We utilize differential coding to effectively reduce the differences in data range and reveal the underlying temporal variations in real-world flight trajectories. Our empirical studies show that using differential values of longitude and latitude can greatly improve prediction accuracy.
    \item We propose FlightPatchNet to fully explore underlying multi-scale temporal patterns. A multi-scale patch network is designed to capture inter- and intra-patch dependencies under different time scales, and integrate multi-scale temporal features across scales and variables. 
    \item We conduct extensive experiments on a real-world dataset. The experiment results demonstrate that our proposed model significantly outperforms the most competitive baselines.
\end{itemize}
  
\section{Related Work}
\textbf{Kinetics-and-Aerodynamics Methods}
The Kinetics-and-Aerodynamics methods \citep{thipphavong2013adaptive,benavides2014implementation,soler2015multiphase, tang20154d} divide the entire flight process into several phases, and establish motion equations for each phase to formulate the flight status. For example, \citet{wang2009prediction} adopted basic flight models to construct horizontal, vertical, and velocity profiles based on the characteristics of different flight phases. \citet{Zhijing7867472} combined the dynamics-and-kinematics models and grayscale theory to predict future trajectories. The grayscale theory can address the parameter missing problem in dynamics-and-kinematics models and improve the prediction performance. Due to numerous unknown and time-varying flight parameters of aircraft, these fixed-parameter methods cannot accurately describe the flight status, leading to poor performance and limited application scenarios.

\textbf{State-Estimation Methods}
The Kalman Filter and its variants \citep{xi2008simulation, Yan6972562} are the typical single-model state-estimation algorithms for FTP tasks, which applies the predefined state equations to estimate the next flight status based on the current observation. For example, \citet{xi2008simulation} applied the Kalman Filter to track discrete flight trajectories by calculating a continuous state transition matrix. However, single-model algorithms cannot adapt to the complex ATC environment. To address this issue, Interactive Multi-Model algorithms \citep{hwang2003flight, li2005survey} have been proposed and successfully applied for trajectory analysis. Although multi-model algorithms can achieve better prediction performance, the computational complexity is high and cannot satisfy the real-time requirement. 

\textbf{Deep Learning Methods}
With the rapid development of deep learning, there has been a surge of deep learning methods for FTP task \citep{ xu2021multi, pang2022bayesian, Sahadevan, Zhang2023FlightTP, Guo2023FlightBERT, Guo2023FlightBERT++}. These learning-based approaches can extract high-dimensional features from raw data, which have achieved a more magnificent performance compared to previous methods. For example, \citet{Sahadevan} used a Bi-directional Long-Short-Term-Memory (Bi-LSTM) network to explore both forward and backward dependencies in the sequential trajectory data. \citet{Zhang2023FlightTP} proposed a wavelet transform-based framework (WTFTP) to perform time-frequency analysis of flight patterns for trajectory prediction.
FlightBERT \citep{Guo2023FlightBERT} employed binary encoding to represent the attributes of the trajectory points and considered the FTP task as a multi-binary classification problem. However, these works predict the future trajectory recursively and suffer from serious error accumulation. Recently, FlightBERT++ \citep{Guo2023FlightBERT++} has been introduced for DMS prediction, which considers the prior horizon information and directly predicts the differential values between adjacent points. 
\section{Methodology}
The short-term FTP task can be formulated as a Multivariate Time Series (MTS) forecasting problem. Given a sequence of historical observations $\mathbf{X} = \left\{\mathbf{x}_1,...,\mathbf{x}_L\right\} \in \mathbb{R}^{C \times L}$,
where  $C$ is the state dimension, $L$ is the look-back window size and $\mathbf{x}_t \in \mathbb{R}^{C \times 1} $  is the flight state at time step $t$. The task is to predict future $T$ time steps  $\hat{\mathbf{Y}} = \left\{\hat{\mathbf{x}}_{L+1},...,\hat{\mathbf{x}}_{L+T}\right\} \in \mathbb{R}^{C' \times T}$, where $C'$ is the predicted state dimension. In this work, the flight state  $\mathbf{x}_t$ represents longitude, latitude, altitude, and velocities along the previous three dimensions, i.e., $\mathbf{x}_t =(Lon_t, Lat_t, Alt_t, Vx_t, Vy_t, Vz_t)^\top$.

\begin{figure*}[h]
    \centering
    \includegraphics[width=0.85\linewidth]{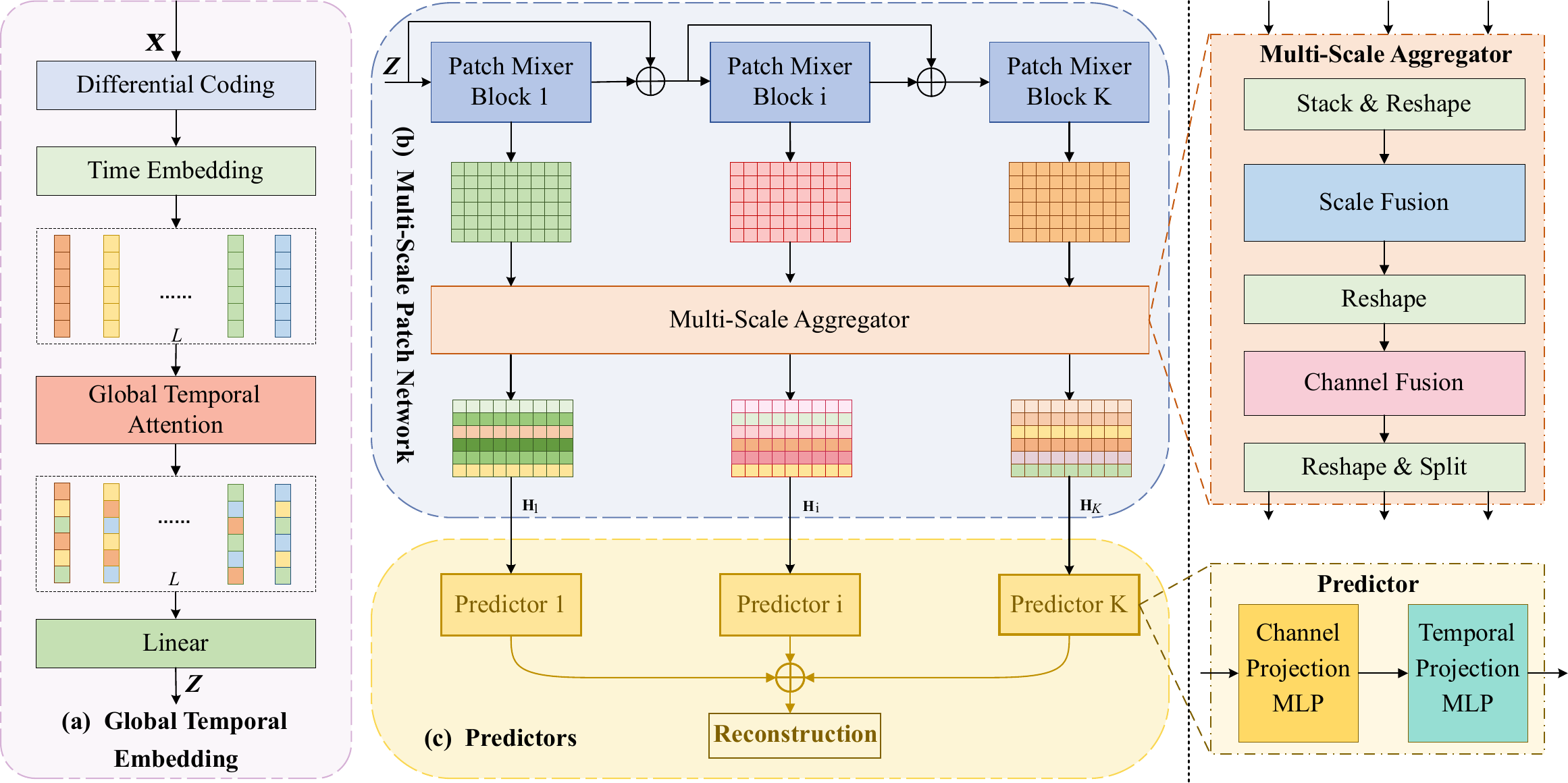}
    \captionsetup{font=small}
    \caption{FlightPatchNet architecture. (a) Global Temporal Embedding to explore the correlations between different time steps. (b) Multi-Scale Patch Network to capture inter- and intra-patch dependencies under different time scales and integrate temporal features across scales and variables. (c) Predictors to exploit complementary temporal features and make direct multi-step prediction. }
    \label{fig:model}
\end{figure*}

The overall architecture of \textbf{FlightPatchNet} is shown in Figure~\ref{fig:model}, which consists of Global Temporal Embedding, Multi-Scale Patch Network, and Predictors. \textbf{Global Temporal Embedding} first utilizes differential coding to transform the original values of longitude and latitude into first-order differences and embeds all variables of the same time step into temporal tokens. Global temporal attention is then introduced to capture the inherent dependencies between different tokens. \textbf{Multi-Scale Patch Network} is proposed to serve as the backbone which is composed of stacked patch mixer blocks and a multi-scale aggregator. Stacked patch mixer blocks divide trajectory series into patches of different sizes from large scale to small scale. Based on divided patches, each patch mixer block exploits a patch encoder and decoder to capture inter- and intra-patch dependencies, endowing our model with powerful temporal modeling capability. To further integrate multi-scale temporal patterns, a multi-scale aggregator is incorporated into the network to capture scale-wise correlations and inter-variable relationships. \textbf{Predictors} provide direct multi-step trajectory forecasting and each predictor is a fully connected network. All the predictor results are aggregated to reconstruct the final prediction trajectory. 
 
\subsection{Global Temporal Embedding}
\paragraph{Differential Coding} In the context of the WGS84 Coordinate System, the longitude and latitude are limited to the intervals $[-180^\circ, 180^\circ]$ and $[-90^\circ, 90^\circ]$ respectively, while the altitude can span from 0 up to tens of thousands of meters. The significant differences of data range caused by physical units may impair the trajectory prediction performance. Generally, normalization algorithms are applied to address this issue. However, the normalized prediction errors should be transformed into raw data range to evaluate the actual performance. For example, if the absolute prediction error of longitude is $10^{-4}$ after using the Min-Max normalization algorithm, the actual prediction error is 0.036° (approximately 4000 meters). Moreover, as shown in Figure 1, the original series of longitude and latitude are over-smoothing and reflect the overall flight trend over a period. If temporal patterns are learned from the original values of longitude and latitude, the model fails to explore the implicit semantic information and cannot focus on short-term temporal variations in flight trajectories.

To address the above issues, we utilize first-order differences for longitude and latitude while original values for other variables, then the differential values are transformed into meters. This process can be formulated as:
\begin{equation}\label{equ:differential}
\left\{
    \scalebox{0.9}{$\begin{aligned}
      \Delta_{\mathit{Lon}} &= 2R\times arcsin(\sqrt{cos^2(\varphi_{t-1})sin^2(\frac{\phi_{t}-\phi_{t-1}}{2}}))  \\
      \Delta_{\mathit{Lat}} &= 2R \times arcsin(\sqrt{sin^2(\frac{\varphi_{t}-\varphi_{t-1}}{2})}) \\
    \end{aligned}$}
    \right.
\end{equation}
where $\phi$ denotes the longitude,  $\varphi$ denotes the latitude, and $R$ is the radius of the earth. By using differential coding for longitude and latitude, the differences in data range are effectively reduced. For example, in our dataset, the range of latitude in original data is about $[-46^\circ, 70^\circ]$ and that in differential data is about $[-3860m, 3860m]$, which spans a similar data range as the altitude. Compared to the original sequences, the differential series can explicitly reflect the underlying temporal variations, which is essential for short-term temporal modeling. Besides, adopting the first-order differences instead of second- or higher-order differences enables the model to reconstruct the predicted trajectory based on the last observation. Note that
we utilize the original values of altitude as inputs rather than differential values. One important reason is that altitude is more susceptible to noise, failing to reflect the actual temporal variations. To this end, the flight state at time step $t$  becomes $ \mathbf{x}_t=(\Delta_{\mathit{Lon}},\Delta_{\mathit{Lat}}, Alt_t, Vx_t, Vy_t, Vz_t)^\top$.

\paragraph{Global Temporal Attention} 
Given the trajectory series $\mathbf{X} \in \mathbb{R}^{C \times L} $,  we first project flight state at each time step into $d$ dimension to generate temporal embeddings $ \mathbf{T}^0  \in \mathbb{R}^{L\times d}$. Then, we apply multi-head self-attention (MSA) \citep{Vaswani2017AttentionIA} on the dimension $L$ to capture the dependencies across all time steps. After attention, the embedding at each time step is enriched with temporal information from other time steps.  This process is formulated as:
\begin{equation}
    \scalebox{0.9}{$
      \begin{aligned}
      \mathbf{T}^0 &= \mathit{TimeEmbedding}(\mathbf{X}^\top) \\
       \mathbf{T}^{i}&=\mathit{LayerNorm}(\mathbf{T}^{i-1}+\mathit{MSA}(\mathbf{T}^{i-1}), i=1,\dots,l \\    \mathbf{T}^{i}&=\mathit{LayerNorm}(\mathbf{T}^{i}+FC(\mathbf{T}^{i}), i=1,\dots,l   \\
       \mathbf{Z} &={(\mathit{Linear}(\mathbf{T}^{l}}))^\top
      \end{aligned}$}
  \end{equation}
 where $l$ is the number of attention layers, $LayerNorm$ denotes the layer normalization \citep{ba2016layer} which has been widely adopted to address non-stationary issues, $MSA$ is the multi-head self-attention layer, $FC$ denotes a fully-connected layer and  $Linear$ projects the embedding of each time step to dimension $C$, i.e., $\mathbb{R}^{d} \rightarrow \mathbb{R}^C$. 

\subsection{Multi-Scale Patch Network}
Considering different temporal patterns prefer diverse time scales, the multi-scale patch network first utilizes a stack of $K$ patch mixer blocks to capture underlying temporal patterns from large scale to small scale. A large time scale can reflect the slow-varying flight trends, while a smaller scale can retain fine-grained local details. To further promote the collaboration of diverse temporal features, a multi-scale aggregator is introduced to consider the contributed scales and dominant variables. Such a multi-scale network equips our model with powerful and complete temporal modeling capability and helps preserve all kinds of multi-scale characteristics.

\subsubsection{Patch Mixer Block}
\paragraph{Patching} Only considering one single time step is insufficient for the FTP task, since it contains limited semantic information and cannot accurately reflect the flight trajectory variations. Inspired by PatchTST \citep{Yuqietal-2023-PatchTST}, the trajectory representation $\mathbf{Z} \in\mathbb{R}^{C\times L}$ is segmented into several non-overlapping patches along the temporal dimension, generating a sequence of patches $\mathbf{Z}_p\in \mathbb{R}^{C\times P\times N}$, where $P$ is the length of each patch, $N$ represents the number of patches, and $N=\left\lceil\frac{L}{P}\right\rceil$. The patching process is formulated as:
\begin{equation}
 {\mathbf{Z}}_p =  {Reshape}({ZeroPadding}(\mathbf{Z})) 
\end{equation}
where $ZeroPadding(\cdot)$ refers to padding series with zeros in the beginning to ensure the length is divisible by $P$.

\paragraph{Patch Encoder-Decoder}
Based on the divided patches $\mathbf{Z}_p $, we utilize a patch encoder and decoder to capture temporal features in flight trajectories. Specifically, the patch encoder aims to capture the inter-patch features (i.e., the global correlations across patches) and intra-patch features (i.e., the local details within patches). After that, these features are reconstructed to the original dimension by the patch decoder. Due to the superiority of linear models for MTS \citep{chen2023tsmixer, zeng2023transformers}, the patch encoders and decoders are based on pure multi-layer perceptron (MLP) for temporal modeling.

\begin{figure}[htbp]
    \centering    
    \includegraphics[width=0.98\linewidth]{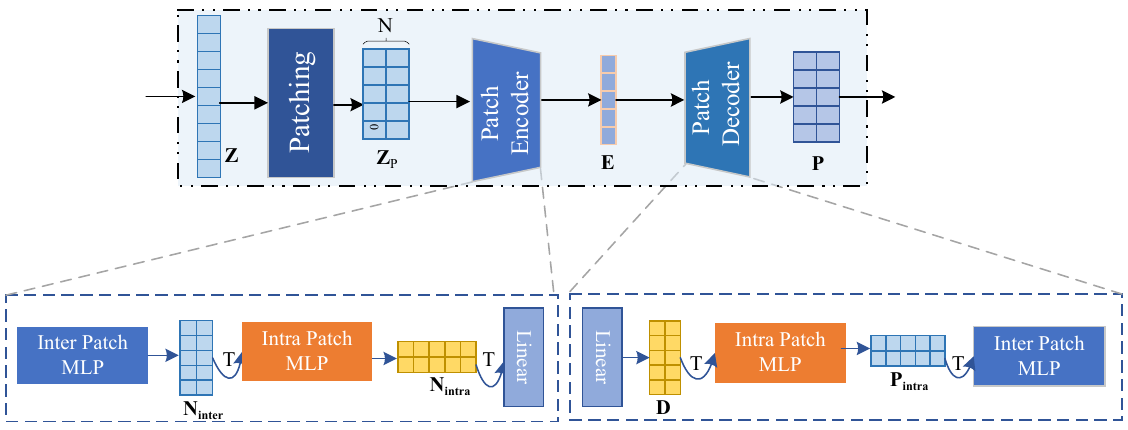}
    \captionsetup{font=small}
    \caption{The structure of Patch Mixer Block}   
    \label{fig:patchmixer}   
\end{figure}
As illustrated in Figure~\ref{fig:patchmixer}, a patch encoder consists of an inter-patch MLP, an intra-patch MLP, and a linear projection. Each MLP has two fully connected layers, a GELU non-linearity layer and a dropout layer with a residual connection.

Given the patch-divided series $\mathbf{Z}_p$, an  inter-patch MLP performs on the dimension $N$ to  capture the dependencies between different patches, which maps $ \mathbb{R}^{N} \rightarrow \mathbb{R}^{N}$ to obtain the inter-patch mixed representation $ \mathbf{N}_{inter} \in \mathbb{R}^{C \times P\times N}$:
\begin{equation}
     \mathbf{N}_{inter} = \mathbf{Z}_{p} + Dropout(FC(\sigma(FC( \mathbf{Z}_p)))) 
\end{equation}
where $\sigma$ denotes a GELU non-linearity layer, $Dropout$  denotes a dropout layer and $\mathbf{N}_{inter}$ reflects the global correlations across patches. After that, an intra-patch MLP  performs on the dimension $P$  to capture the dependencies  across different time steps within patches, which maps $ \mathbb{R}^{P} \rightarrow \mathbb{R}^{P}$ to obtain  the intra-patch mixed representation $ \mathbf{N}_{intra} \in \mathbb{R}^{C \times N\times P}$:
\begin{equation}
    \mathbf{N}_{intra} = \mathbf{N}_{inter}^\top + Dropout(FC(\sigma(FC( \mathbf{N}_{inter}^\top))))
\end{equation} where $ \mathbf{N}_{intra} $ reflects the local details between different time steps within patches. Then, we perform a linear projection on  $\mathbf{N}_{intra}^\top$ to obtain the final inter- and intra-patch mixed  representation $\mathbf{E}$ $\in \mathbb{R}^{C \times P\times 1}$:
\begin{equation}
     \mathbf{E} = \mathit{Linear}( \mathbf{N}_{intra}^\top)
\end{equation}
After such a patch encoding process, the correlations between nearby time steps within patches and distant time steps across patches are finely explored. Then, we utilize a patch decoder to reconstruct the original sequence. A patch decoder comprises the same components as the encoder in a reverse order, which is formulated as follows:
\begin{equation}
    \begin{aligned}
 \mathbf{D} &= Linear( \mathbf{E}) \\
 \mathbf{P}_{intra} &= \mathbf{D}^\top + Dropout(FC(\sigma(FC(\mathbf{D}^\top))))\\
  \mathbf{P} &= \mathbf{P}_{intra}^\top + Dropout(FC(\sigma(FC(\mathbf{P}_{intra}^\top))))
    \end{aligned}
\end{equation}
where $Linear$ makes a dimensional projection to obtain $ \mathbf{D} \in \mathbb{R}^{C \times P\times N}$ for reconstructing the original sequence, $\mathbf{P}_{intra} \in \mathbb{R}^{C \times N\times P}$ is the reconstructed intra-patch mixed representation, and $\mathbf{P} \in \mathbb{R}^{C \times P\times N} $ is the final reconstructed intra- and inter-patch mixed representation.

\subsubsection{Multi-Scale Aggregator} 
To enable the ability of more complete multi-scale modeling, we introduce a multi-scale aggregator to integrate different temporal patterns. It contains two components: scale fusion and channel fusion. Scale fusion can figure out critical time scales and capture the scale-wise correlations, while channel fusion can discover dominant variables affecting temporal variations and explore the inter-variable relationships. These two components work together to help the model learn a robust multi-scale representation and improve generalization ability. 
Given the $K$ scale-specific temporal representations $\{\mathbf{P}_1,\mathbf{P}_2,\dots,\mathbf{P}_K\}$, we first stack them and rearrange the data to combine the three dimensions of channel size $C$, patch size $P$ and patch quantity $N$, resulting in $\mathbf{S}^0 \in \mathbb{R}^{K\times (C\times L)}$, where $L = P \times N$. Then we apply MSA on the scale dimension $K$ to learn the importance of contributed time scales. This process is formulated as:
\begin{equation}
      \begin{aligned}
      \mathbf{S}^0&= Reshape(Stack(\mathbf{P}_1,\mathbf{P}_2,\dots,\mathbf{P}_K) ) \\
       \mathbf{S}^{i}&=LayerNorm(\mathbf{S}^{i-1}+MSA(\mathbf{S}^{i-1})\\     
       \mathbf{S}^{i}&=LayerNorm(\mathbf{S}^{i}+FC(\mathbf{S}^{i}), i=1,\dots,l    
      \end{aligned}
  \end{equation}
where $\mathbf{S}^l$ is the final multi-scale fusion representation within variables.
Inspired by iTransformer \citep{liu2023itransformer}, we consider each variable as a token and apply MSA to explore dependencies between different variables. We first reshape the  $\mathbf{S}^l$ to get  $\mathbf{C}^0$ $ \in \mathbb{R}^{C \times (K\times L)}$ and perform multi-head self-attention on the channel dimension $C$ to identify dominant variables. This process is simply formulated as follows:
\begin{equation}
    \begin{aligned}
        \mathbf{C}^0 &= Reshape(\mathbf{S}^l) \\
        \mathbf{C}^{i} &=LayerNorm(\mathbf{C}^{i-1}+ \mathit{MSA}(\mathbf{C}^{i-1})) \\
         \mathbf{C}^{i}&=\mathit{LayerNorm}(\mathbf{C}^{i}+FC(\mathbf{C}^{i}),i=1,\dots,l \\
         \mathbf{H} &= Reshape(\mathbf{C}^l) 
    \end{aligned}
\end{equation} where $\mathbf{H} \in \mathbb{R}^{ C  \times L \times K}$ is the final multi-scale representation which involves cross-scale correlations and inter-variable relationships.
\subsection{Direct Multi-Step Prediction}
We ensemble $K$ predictors to directly obtain the future flight trajectory series, which can exploit complementary information from different temporal patterns. The objective of our model is to predict the differential values of longitude and latitude relative to the last observation, and the raw absolute values of altitude, i.e., $\hat{\mathbf{Y}} = \left\{\hat{\mathbf{x}}_{L+1},...,\hat{\mathbf{x}}_{L+T}\right\}$, where 
$ \hat{\mathbf{x}}_{L+i} =( \hat{\Delta}^{\mathit{Lon}}(L+i,L) ,  \hat{\Delta}^{\mathit{Lat}}(L+i,L),\hat{Alt}_{L+i})^\top$ for $i=1,\dots,T$. 
We split the final multi-scale representation $\mathbf{H} \in \mathbb{R}^{ C  \times L \times K}$ into a sequence $\left\{ \mathbf{H}_{*, 1},\mathbf{H}_{*, 2},\dots,\mathbf{H}_{*, K}\right\}$, where $\mathbf{H}_{*, i}\in \mathbb{R}^{ C  \times L}$ for $i=1,\dots,K$, and feed each $\mathbf{H}_{*, i}$ to a predictor. Each predictor has two MLPs. The first $MLP_{C_i}$ transforms the input channel $C$ into the output channel $C'$, and the second $MLP_{T_i}$ projects the historical input sequence $L$ to the prediction horizon $T$. 
\begin{equation}
    \begin{aligned}
        \hat{\mathbf{Y}_i} = &MLP_{T_i}(MLP_{C_i}(\mathbf{H}_{*, i})) \\
\hat{\mathbf{Y}}=&\sum\limits_{i=1}^K \hat{\mathbf{Y}_i}
    \end{aligned}
\end{equation}
Finally, all the predictor results are aggregated to reconstruct the final prediction trajectory according to Equation (\ref{equ:differential}), which can enhance the stability and generalization of our model.

\section{Experiments}\label{main:exp}
\subsection{Dataset and Experimental Setup}
\paragraph{Datasets} To evaluate the performance of FlightPatchNet, we conduct extensive experiments on ADS-B data provided by OpenSky \footnote[1]{https://opensky-network.org/datasets/states/} from 2020 to 2022. In this paper, six key attributes are extracted from the original data, including longitude, latitude, altitude, and velocity in \textit{x, y, z} dimensions. The dataset is chronologically divided into three parts for training, validation, and testing with a ratio of 8:1:1. 
\paragraph{Baselines and Setup} We compare our model with ten competitive models, including five IMS-based models: \textbf{LSTM} \citep{LSTM8489734}, \textbf{CNN-LSTM} \citep{CNN-LSTM9145522}, \textbf{Bi-LSTM} \citep{Sahadevan}, \textbf{FlightBERT} \citep{Guo2023FlightBERT}, \textbf{WTFTP} \citep{Zhang2023FlightTP};  five DMS-based model: \textbf{FlightBERT++} \citep{Guo2023FlightBERT++}, \textbf{TimeMixer} \citep{wang2024timemixer}, \textbf{TimesNet} \citep{wu2022timesnet}, \textbf{MICN} \citep{wang2023micn}, \textbf{Pathformer} \citep{chen2024pathformer}. These models have covered mainstream deep learning architectures, including Transformer (FlightBERT, FlightBERT++, Pathformer), CNN (CNN-LSTM, TimesNet, MICN), RNN (LSTM, Bi-LSTM, CNN-LSTM, WTFTP) and MLP (TimeMixer), which helps to provide a comprehensive comparison. For fairness, all the models follow the same experimental setup with lookback window $L=60$ and prediction horizon  $T\in\{1,3,9,15\}$.  Our model is trained with MSE loss, using the Adam optimizer. We adopt the Mean Absolute Error (MAE) and Root Mean Squared Error (RMSE)  as evaluation metrics. More details about the dataset, baselines, implementation and hyper-parameters are shown in Appendix~\ref{app:exp}.
\subsection{Main Results}
Comprehensive flight prediction results are demonstrated in Table~\ref{table:exp_main_res} (see Appendix~\ref{error_bar} for error bar). FlightPatchNet achieves the most outstanding performance across various prediction lengths for longitude and latitude in terms of both MAE and RMSE, while it does not achieve the optimal results for altitude compared with other strong baselines such as FlightBERT++. For simplification, we consider prediction horizon $T=15$ and compare our model with the second best. FlightPatchNet achieves an overall 18\textit{.}62\% reduction on MAE and 41\textit{.}29\% reduction on RMSE for longitude, and 35\textit{.}31\% reduction on MAE and 44\textit{.}80\% reduction on RMSE for latitude. For the prediction performance of altitude, FlightBERT++ outperforms our model by 45\textit{.}51 meters reduction on MAE but has a large  RMSE which may caused by high-bit errors in the prediction. FlightPatchNet obtains the smallest RMSE results for all variables, indicating that our model can provide a more robust and stable prediction. Compared with the most promising multi-scale modeling MTS prediction methods, including the pure MLP-based model TimeMixer, the CNN-based methods TimesNet and MICN and the Transformer-based method Pathformer, FlightPatchNet achieves superior prediction performance. These existing multivariate time-series forecasting methods typically decompose time series into seasonal and trend components, and primarily focus on periodic modeling. However, short-term flight trajectories do not exhibit obvious periodic patterns. The trend features in altitude and the temporal variations in longitude and latitude deserve more attention. Furthermore, as the prediction horizon increases, IMS-based models suffer from serious performance degradation due to error accumulation. 

\begin{table*}[ht]
\centering
\captionsetup{font=small}
\caption{Flight trajectory prediction results.  A lower MAE or RMSE represents a better prediction. The prediction horizon \textit{$T \in \left\{1,3,9,15\right\}$} and look-back window size $L = 60$ for all experiments. The best results are highlighted in \textbf{bold }and the second best are \underline{underlined}. Note that $0.00001^\circ$ is about 1m.}
\label{table:exp_main_res}

\setlength{\tabcolsep}{1mm}{
\scalebox{0.7}{
    \begin{tabular}{c|c|c|cccc|cccc|cccc}
\toprule
\multirow{2}{*}{}     & \multirow{2}{*}{Model}                                                           & \multirow{2}{*}{Metric} & \multicolumn{4}{c|}{Lon($0.00001^\circ$)}                                                                                                                                                 & \multicolumn{4}{c|}{Lat($0.00001^\circ$)}                                                                                                                                                & \multicolumn{4}{c}{Alt(m)}                                                                                                                                                                           \\ \cmidrule(l){4-15} 
                      &                                                                                  &                         & \multicolumn{1}{c|}{1}                           & \multicolumn{1}{c|}{3}                            & \multicolumn{1}{c|}{9}                             & 15                            & \multicolumn{1}{c|}{1}                           & \multicolumn{1}{c|}{3}                            & \multicolumn{1}{c|}{9}                            & 15                            & \multicolumn{1}{c|}{1}                               & \multicolumn{1}{c|}{3}                               & \multicolumn{1}{c|}{9}                               & 15                              \\ \midrule
\multirow{10}{*}{IMS} & \multirow{2}{*}{LSTM}                                                            & MAE                     & \multicolumn{1}{c|}{56} & \multicolumn{1}{c|}{427}                          & \multicolumn{1}{c|}{1747}                          & 3132                          & \multicolumn{1}{c|}{49} & \multicolumn{1}{c|}{493}                          & \multicolumn{1}{c|}{2116}                         & 3717                          & \multicolumn{1}{c|}{92.27}                           & \multicolumn{1}{c|}{159.30}                          & \multicolumn{1}{c|}{549.55}                          & 882.86                          \\ \cmidrule(l){3-15} 
                      &                                                                                  & RMSE                    & \multicolumn{1}{c|}{95} & \multicolumn{1}{c|}{691}                          & \multicolumn{1}{c|}{2597}                          & 4578                          & \multicolumn{1}{c|}{89} & \multicolumn{1}{c|}{740}                          & \multicolumn{1}{c|}{2956}                         & 5143                          & \multicolumn{1}{c|}{142.05}                          & \multicolumn{1}{c|}{233.39}                          & \multicolumn{1}{c|}{763.84}                          & 768.45                          \\ \cmidrule(l){2-15} 
                      & \multirow{2}{*}{Bi-LSTM}                                                         & MAE                     & \multicolumn{1}{c|}{155}                         & \multicolumn{1}{c|}{747}                          & \multicolumn{1}{c|}{2319}                          & 3890                          & \multicolumn{1}{c|}{137}                         & \multicolumn{1}{c|}{824}                          & \multicolumn{1}{c|}{2711}                         & 4404                          & \multicolumn{1}{c|}{432.50}                          & \multicolumn{1}{c|}{761.50}                          & \multicolumn{1}{c|}{1648.68}                         & 2006.21                         \\ \cmidrule(l){3-15} 
                      &                                                                                  & RMSE                    & \multicolumn{1}{c|}{202}                         & \multicolumn{1}{c|}{1124}                         & \multicolumn{1}{c|}{3387}                          & 5532                          & \multicolumn{1}{c|}{181}                         & \multicolumn{1}{c|}{1142}                         & \multicolumn{1}{c|}{3639}                         & 5982                          & \multicolumn{1}{c|}{563.74}                          & \multicolumn{1}{c|}{953.37}                          & \multicolumn{1}{c|}{2132.91}                         & 2420.74                         \\ \cmidrule(l){2-15} 
                      & \multirow{2}{*}{CNN-LSTM}                                                        & MAE                     & \multicolumn{1}{c|}{139}                         & \multicolumn{1}{c|}{700}                          & \multicolumn{1}{c|}{2282}                          & 4149                          & \multicolumn{1}{c|}{131}                         & \multicolumn{1}{c|}{801}                          & \multicolumn{1}{c|}{2623}                         & 5139                          & \multicolumn{1}{c|}{520.03}                          & \multicolumn{1}{c|}{746.67}                          & \multicolumn{1}{c|}{1569.68}                         & 1136.80                         \\ \cmidrule(l){3-15} 
                      &                                                                                  & RMSE                    & \multicolumn{1}{c|}{240}                         & \multicolumn{1}{c|}{1033}                         & \multicolumn{1}{c|}{3263}                          & 5981                          & \multicolumn{1}{c|}{212}                         & \multicolumn{1}{c|}{1130}                         & \multicolumn{1}{c|}{3559}                         & 7353                          & \multicolumn{1}{c|}{1176.96}                         & \multicolumn{1}{c|}{926.40}                          & \multicolumn{1}{c|}{1936.63}                         & 1658.53                         \\ \cmidrule(l){2-15} 
                      & \multirow{2}{*}{WTFTP}                                                           & MAE                     & \multicolumn{1}{c|}{175}                         & \multicolumn{1}{c|}{1484}                         & \multicolumn{1}{c|}{2002}                          & 2657                          & \multicolumn{1}{c|}{112}                         & \multicolumn{1}{c|}{1169}                         & \multicolumn{1}{c|}{1586}                         & 2110                          & \multicolumn{1}{c|}{145.02}                          & \multicolumn{1}{c|}{230.49}                          & \multicolumn{1}{c|}{588.44}                          & 957.41                          \\ \cmidrule(l){3-15} 
                      &                                                                                  & RMSE                    & \multicolumn{1}{c|}{218}                         & \multicolumn{1}{c|}{1905}                         & \multicolumn{1}{c|}{2606}                          & 3531                          & \multicolumn{1}{c|}{171}                         & \multicolumn{1}{c|}{1739}                         & \multicolumn{1}{c|}{2328}                         & 3124                          & \multicolumn{1}{c|}{415.13}                          & \multicolumn{1}{c|}{497.24}                          & \multicolumn{1}{c|}{1021.52}                         & 1583.38                         \\ \cmidrule(l){2-15} 
                      & \multirow{2}{*}{FlightBERT}                                                      & MAE                     & \multicolumn{1}{c|}{123}                         & \multicolumn{1}{c|}{241} & \multicolumn{1}{c|}{1162}                          & 2407                          & \multicolumn{1}{c|}{88}                          & \multicolumn{1}{c|}{\underline{158}} & \multicolumn{1}{c|}{963}                          & 1238                          & \multicolumn{1}{c|}{24.67}                           & \multicolumn{1}{c|}{35.67}                           & \multicolumn{1}{c|}{78.58}                           & 134.29                          \\ \cmidrule(l){3-15} 
                      &                                                                                  & RMSE                    & \multicolumn{1}{c|}{241}                         & \multicolumn{1}{c|}{526} & \multicolumn{1}{c|}{2189}                          & 3969                          & \multicolumn{1}{c|}{154}                         & \multicolumn{1}{c|}{\underline{286}} & \multicolumn{1}{c|}{1904}                         & 3093                          & \multicolumn{1}{c|}{234.17}                          & \multicolumn{1}{c|}{272.59}                          & \multicolumn{1}{c|}{384.22}                          & 462.28                          \\ \midrule
\multirow{12}{*}{DMS} & \multirow{2}{*}{FlightBERT++}                                                    & MAE                     & \multicolumn{1}{c|}{173}                         & \multicolumn{1}{c|}{317}                          & \multicolumn{1}{c|}{\underline{871}}  & \underline{1187} & \multicolumn{1}{c|}{85}                          & \multicolumn{1}{c|}{210}                          & \multicolumn{1}{c|}{\underline{612}} & \underline{1048} & \multicolumn{1}{c|}{\textbf{9.39}}                   & \multicolumn{1}{c|}{\textbf{21.89}}                  & \multicolumn{1}{c|}{\textbf{47.84}}                  & \textbf{78.46}                  \\ \cmidrule(l){3-15} 
                      &                                                                                  & RMSE                    & \multicolumn{1}{c|}{360}                         & \multicolumn{1}{c|}{659}                          & \multicolumn{1}{c|}{\underline{1846}} & 3131 & \multicolumn{1}{c|}{148}                         & \multicolumn{1}{c|}{425}                          & \multicolumn{1}{c|}{\underline{959}} & \underline{2127} & \multicolumn{1}{c|}{175.29} & \multicolumn{1}{c|}{167.16} & \multicolumn{1}{c|}{327.93} & 384.18 \\ \cmidrule(l){2-15} 
                      & \multirow{2}{*}{TimeMixer}                                                       & MAE                     & \multicolumn{1}{c|}{67}                          & \multicolumn{1}{c|}{765}                          & \multicolumn{1}{c|}{3100}                          & 5581                          & \multicolumn{1}{c|}{42}                          & \multicolumn{1}{c|}{422}                          & \multicolumn{1}{c|}{1679}                         & 3028                          & \multicolumn{1}{c|}{21.18}                           & \multicolumn{1}{c|}{50.02}                           & \multicolumn{1}{c|}{119.80}                          & 157.47                          \\ \cmidrule(l){3-15} 
                      &                                                                                  & RMSE                    & \multicolumn{1}{c|}{115}                         & \multicolumn{1}{c|}{1466}                         & \multicolumn{1}{c|}{5517}                          & 9698                          & \multicolumn{1}{c|}{76}                          & \multicolumn{1}{c|}{789}                          & \multicolumn{1}{c|}{2976}                         & 5318                          & \multicolumn{1}{c|}{\textbf{109.57}}                          & \multicolumn{1}{c|}{145.55}                          & \multicolumn{1}{c|}{\underline{281.31}}                          & \underline{374.82}                          \\ \cmidrule(l){2-15} 
                      & \multirow{2}{*}{TimesNet}                                                        & MAE                     & \multicolumn{1}{c|}{73}                          & \multicolumn{1}{c|}{1383}                         & \multicolumn{1}{c|}{5459}                          & 9421                          & \multicolumn{1}{c|}{46}                          & \multicolumn{1}{c|}{753}                          & \multicolumn{1}{c|}{2981}                         & 5086                          & \multicolumn{1}{c|}{36.41}                           & \multicolumn{1}{c|}{79.42}                           & \multicolumn{1}{c|}{178.49}                          & 257.39                          \\ \cmidrule(l){3-15} 
                      &                                                                                  & RMSE                    & \multicolumn{1}{c|}{124}                         & \multicolumn{1}{c|}{2281}                         & \multicolumn{1}{c|}{8377}                          & 14420                         & \multicolumn{1}{c|}{83}                          & \multicolumn{1}{c|}{1237}                         & \multicolumn{1}{c|}{4557}                         & 7796                          & \multicolumn{1}{c|}{154.37}                          & \multicolumn{1}{c|}{184.36}                          & \multicolumn{1}{c|}{392.23}                          & 543.37                          \\ \cmidrule(l){2-15} 
                      & \multirow{2}{*}{MICN}                                                            & MAE                     & \multicolumn{1}{c|}{69}                          & \multicolumn{1}{c|}{680}                          & \multicolumn{1}{c|}{2235}                          & 3912                          & \multicolumn{1}{c|}{\underline{40}}                          & \multicolumn{1}{c|}{384}                          & \multicolumn{1}{c|}{1296}                         & 2256                          & \multicolumn{1}{c|}{48.28}                           & \multicolumn{1}{c|}{544.35}                          & \multicolumn{1}{c|}{1992.86}                         & 3431.19                         \\ \cmidrule(l){3-15} 
                      &                                                                                  & RMSE                    & \multicolumn{1}{c|}{116}                         & \multicolumn{1}{c|}{1133}                         & \multicolumn{1}{c|}{3831}                          & 6598                          & \multicolumn{1}{c|}{\underline{73}}                          & \multicolumn{1}{c|}{633}                          & \multicolumn{1}{c|}{2167}                         & 3704                          & \multicolumn{1}{c|}{\underline{112.31}}                          & \multicolumn{1}{c|}{945.57}                          & \multicolumn{1}{c|}{3270.77}                         & 5612.64                         \\ \cmidrule(l){2-15} 
                      & \multirow{2}{*}{Pathformer}                                                      & MAE                     & \multicolumn{1}{c|}{\underline{52}}                          & \multicolumn{1}{c|}{\underline{232}}                          & \multicolumn{1}{c|}{1374}                          & 1914                          & \multicolumn{1}{c|}{45}                          & \multicolumn{1}{c|}{232}                          & \multicolumn{1}{c|}{863}                          & 1806                          & \multicolumn{1}{c|}{42.03}                           & \multicolumn{1}{c|}{47.65}                           & \multicolumn{1}{c|}{141.08}                          & 259.53                          \\ \cmidrule(l){3-15} 
                      &                                                                                  & RMSE                    & \multicolumn{1}{c|}{\underline{89}}                          & \multicolumn{1}{c|}{\underline{373}}                          & \multicolumn{1}{c|}{2227}                          & \underline{2686}                          & \multicolumn{1}{c|}{83}                          & \multicolumn{1}{c|}{373}                          & \multicolumn{1}{c|}{1346}                         & 2570                          & \multicolumn{1}{c|}{114.37}                          & \multicolumn{1}{c|}{\underline{136.59}}                          & \multicolumn{1}{c|}{406.37}                          & 645.02                          \\ \cmidrule(l){2-15} 
                      & \multirow{2}{*}{\begin{tabular}[c]{@{}c@{}}FlightPatchNet\\ (Ours)\end{tabular}} & MAE                     & \multicolumn{1}{c|}{\textbf{48}}                 & \multicolumn{1}{c|}{\textbf{153}}                 & \multicolumn{1}{c|}{\textbf{546}}                  & \textbf{966}                  & \multicolumn{1}{c|}{\textbf{32}}                 & \multicolumn{1}{c|}{\textbf{105}}                 & \multicolumn{1}{c|}{\textbf{381}}                 & \textbf{678}                  & \multicolumn{1}{c|}{\underline{13.34}}  & \multicolumn{1}{c|}{\underline{32.65}}  & \multicolumn{1}{c|}{\underline{78.57}}  & \underline{123.97} \\ \cmidrule(l){3-15} 
                      &                                                                                  & RMSE                    & \multicolumn{1}{c|}{\textbf{87}}                 & \multicolumn{1}{c|}{\textbf{233}}                 & \multicolumn{1}{c|}{\textbf{885}}                  & \textbf{1577}                 & \multicolumn{1}{c|}{\textbf{64}}                 & \multicolumn{1}{c|}{\textbf{175}}                 & \multicolumn{1}{c|}{\textbf{652}}                 & \textbf{1174}                 & \multicolumn{1}{c|}{123.78}                 & \multicolumn{1}{c|}{\textbf{121.48}}                 & \multicolumn{1}{c|}{\textbf{174.63}}                 & \textbf{244.34}                 \\ \bottomrule
\end{tabular}
}
}
\end{table*}

\begin{figure}[h]
\centering
\subfigure{
    \centering
    \includegraphics[width=0.9\linewidth]{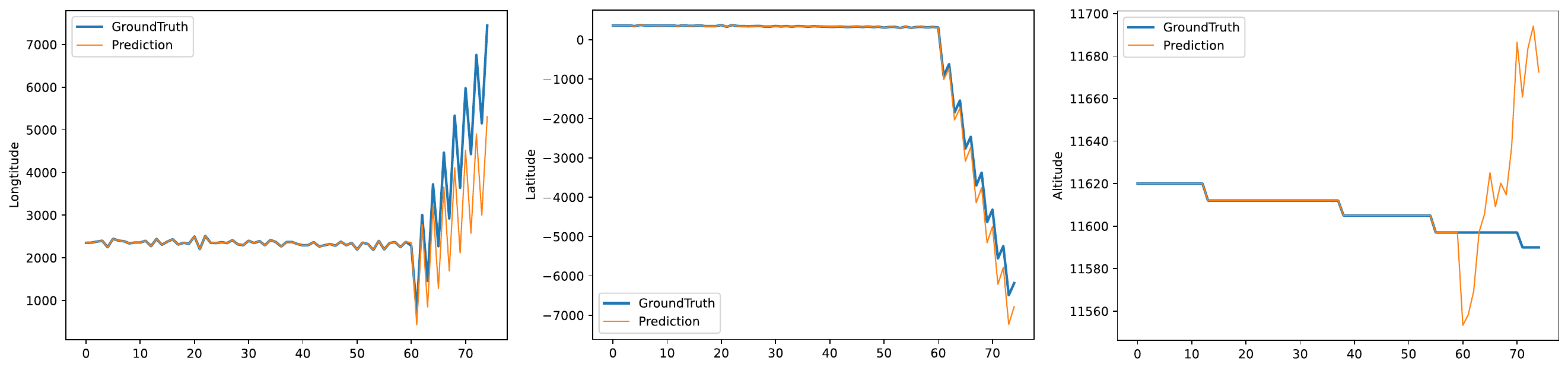}
}\\

\subfigure{
\centering
    \includegraphics[width=0.9\linewidth]{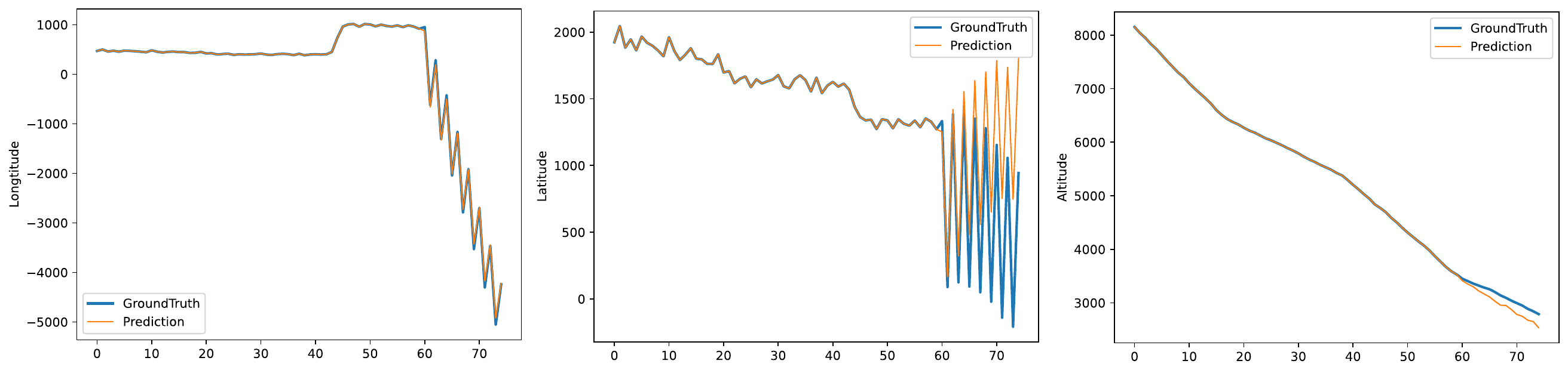}
}\\
\subfigure{
    \centering
    \includegraphics[width=0.9\linewidth]{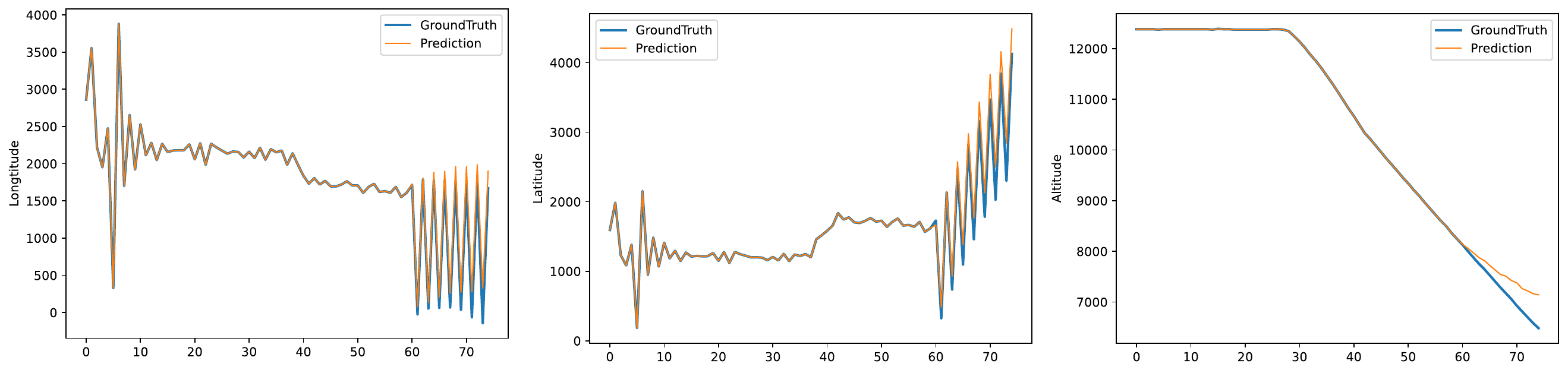}
}
   
    \captionsetup{font=small}
    \caption{Visualization of the ground truth and predictions of FlightPatchNet when the prediction horizon $T=15$ and look-back window size $L=60$. The series of altitude are in original data while those of longitude and latitude are in differential data, all denoted by meters.}
    \label{fig:flightPatchNet_pred}
\end{figure}

\paragraph{Visualization of FlightPatchNet Predictions}  
Figure~\ref{fig:flightPatchNet_pred} shows that FlightPatchNet can comprehensively capture the temporal variations of longitude and latitude, while it fails to fully reveal the temporal patterns from original altitude series. 
\paragraph{Visualization of FlightPatchNet Predictions for Altitude} 
\begin{figure}[h]
\centering
    \subfigure[]{
     \centering
        \includegraphics[width=0.8\linewidth]{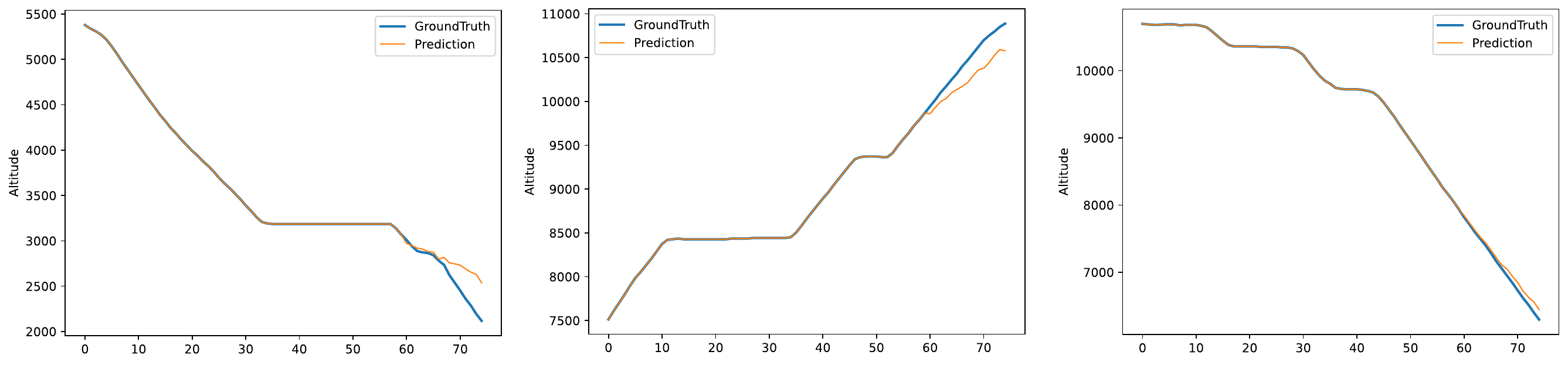}
        \label{altitude_a}
    }
    \subfigure[]{
        \centering
        \includegraphics[width=0.8\linewidth]{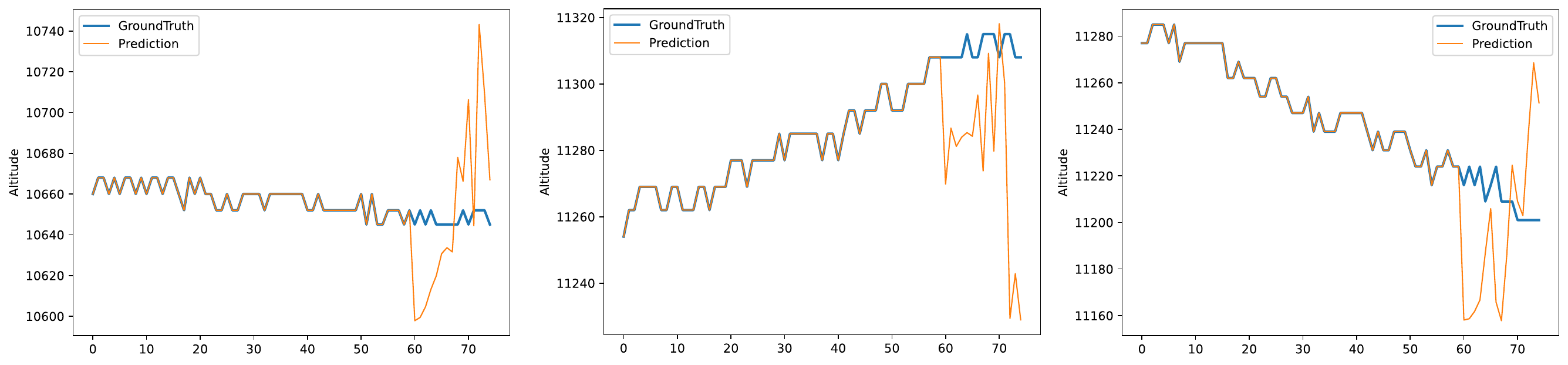}
         \label{altitude_b}
    }\\
    \subfigure[]{
     \centering
        \includegraphics[width=0.8\linewidth]{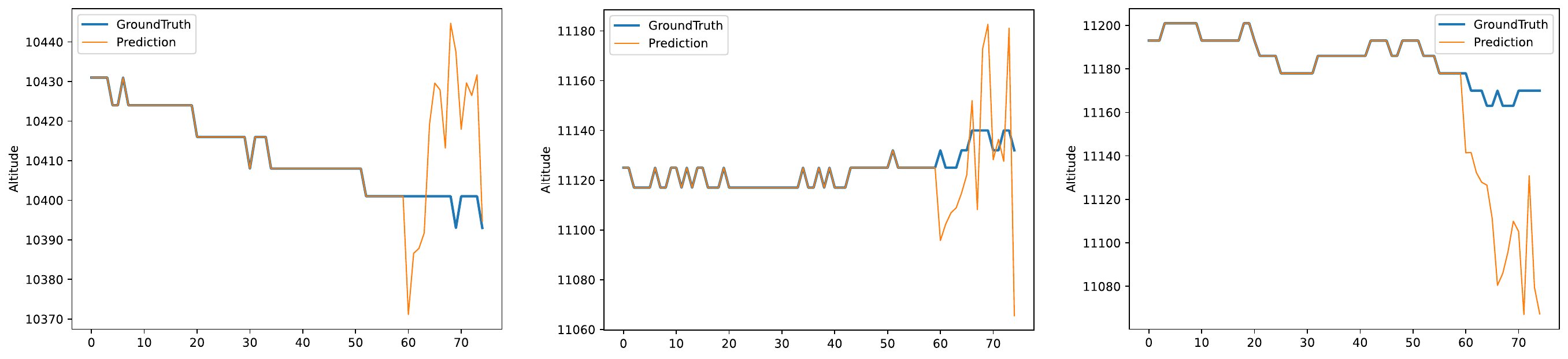}
         \label{altitude_c}
    }
    \captionsetup{font=small}

    \caption{Visualization of ground truth for altitude and predictions of FlightPatchNet when the prediction horizon $T=15$ and look-back window size $L=60$, all denoted by meters.}
    \label{altitude_pred_gt}
\end{figure}
We present the visualization of FlightPatchNet predictions and ground truth for the altitude in Figure~\ref{altitude_pred_gt}. As shown in Figure~\ref{altitude_a}, when the series of altitudes are relatively smooth and stationary with obvious global trends, FlightPatchNet can effectively capture these trends and make accurate predictions. When the series suffers from many change points caused by frequent abrupt fluctuations, as depicted in Figure~\ref{altitude_b} and ~\ref{altitude_c}, FlightPatchNet tends to focus more on the irregular change points during prediction, leading to a large deviation from the ground truth. As a result, FlightPatchNet struggles to capture the real temporal variations in altitude and fails to provide accurate predictions. 

\paragraph{Effectiveness of Differential Coding}
The results in Table~\ref{table:exp_res_diff} show that using differential coding for longitude and latitude can significantly improve their prediction performance but slightly decrease the accuracy of altitude. The differential coding can reveal the temporal variations of longitude and latitude, which helps the temporal modeling in flight trajectories. However, the variations of altitude in the original series may come from unexpected noise. FlightPatchNet has a strong modeling capacity for temporal variations and tends to focus more on the noise points during altitude prediction, leading to a large bias towards the ground truth.

\begin{table}[h]
\centering
\captionsetup{font=small}
\caption{Flight trajectory prediction results for longitude and latitude in original data and differential data when prediction horizon $T=15$. The best results are highlighted in \textbf{bold}. Note that altitude and velocities are always in original data.}
\label{table:exp_res_diff}

\setlength{\tabcolsep}{1mm}{
    \scalebox{0.78}{
        \begin{tabular}{@{}cccccc@{}}
\toprule
Models                                                                                                & Diff                                               & Metric & Lon($^\circ$)      & Lat($^\circ$)      & Alt(m)       \\ \midrule
\multicolumn{1}{c|}{\multirow{4}{*}{LSTM}}                                                            & \multicolumn{1}{c|}{\multirow{2}{*}{$\checkmark$}} & MAE    & 0.03132          & 0.03717          & 883          \\ \cmidrule(l){3-6} 
\multicolumn{1}{c|}{}                                                                                 & \multicolumn{1}{c|}{}                              & RMSE   & 0.04578          & 0.05143          & 1206         \\ \cmidrule(l){2-6} 
\multicolumn{1}{c|}{}                                                                                 & \multicolumn{1}{c|}{\multirow{2}{*}{$\times$}}     & MAE    & 0.82230          & 0.12008          & 769          \\ \cmidrule(l){3-6} 
\multicolumn{1}{c|}{}                                                                                 & \multicolumn{1}{c|}{}                              & RMSE   & 1.20424          & 2.44136          & 1053         \\ \midrule
\multicolumn{1}{c|}{\multirow{4}{*}{Bi-LSTM}}                                                         & \multicolumn{1}{c|}{\multirow{2}{*}{$\checkmark$}} & MAE    & 0.03890          & 0.04404          & 2006         \\ \cmidrule(l){3-6} 
\multicolumn{1}{c|}{}                                                                                 & \multicolumn{1}{c|}{}                              & RMSE   & 0.05532          & 0.05982          & 2421         \\ \cmidrule(l){2-6} 
\multicolumn{1}{c|}{}                                                                                 & \multicolumn{1}{c|}{\multirow{2}{*}{$\times$}}     & MAE    & 1.71433          & 0.19014          & 2091         \\ \cmidrule(l){3-6} 
\multicolumn{1}{c|}{}                                                                                 & \multicolumn{1}{c|}{}                              & RMSE   & 2.43607          & 0.27621          & 2666         \\ \midrule
\multicolumn{1}{c|}{\multirow{4}{*}{CNN-LSTM}}                                                        & \multicolumn{1}{c|}{\multirow{2}{*}{$\checkmark$}} & MAE    & 0.04149          & 0.05139          & 1137         \\ \cmidrule(l){3-6} 
\multicolumn{1}{c|}{}                                                                                 & \multicolumn{1}{c|}{}                              & RMSE   & 0.05981          & 0.07353          & 1659         \\ \cmidrule(l){2-6} 
\multicolumn{1}{c|}{}                                                                                 & \multicolumn{1}{c|}{\multirow{2}{*}{$\times$}}     & MAE    & 8.59512          & 1.95957          & 1638         \\ \cmidrule(l){3-6} 
\multicolumn{1}{c|}{}                                                                                 & \multicolumn{1}{c|}{}                              & RMSE   & 23.07600         & 8.15418          & 2114         \\ \midrule
\multicolumn{1}{c|}{\multirow{4}{*}{\begin{tabular}[c]{@{}c@{}}FlightPatchNet\\ (Ours)\end{tabular}}} & \multicolumn{1}{c|}{\multirow{2}{*}{$\checkmark$}} & MAE    & \textbf{0.00966} & \textbf{0.00678} & 124          \\ \cmidrule(l){3-6} 
\multicolumn{1}{c|}{}                                                                                 & \multicolumn{1}{c|}{}                              & RMSE   & \textbf{0.01577} & \textbf{0.01174} & 244          \\ \cmidrule(l){2-6} 
\multicolumn{1}{c|}{}                                                                                 & \multicolumn{1}{c|}{\multirow{2}{*}{$\times$}}     & MAE    & 0.19348          & 0.05385          & \textbf{61}  \\ \cmidrule(l){3-6} 
\multicolumn{1}{c|}{}                                                                                 & \multicolumn{1}{c|}{}                              & RMSE   & 0.26243          & 0.07457          & \textbf{170} \\ \bottomrule
\end{tabular}
    }
}
\end{table}
\paragraph{Effectiveness of Multi Scales} 

\begin{figure}[h]
     \centering
        \includegraphics[width=0.8\linewidth]{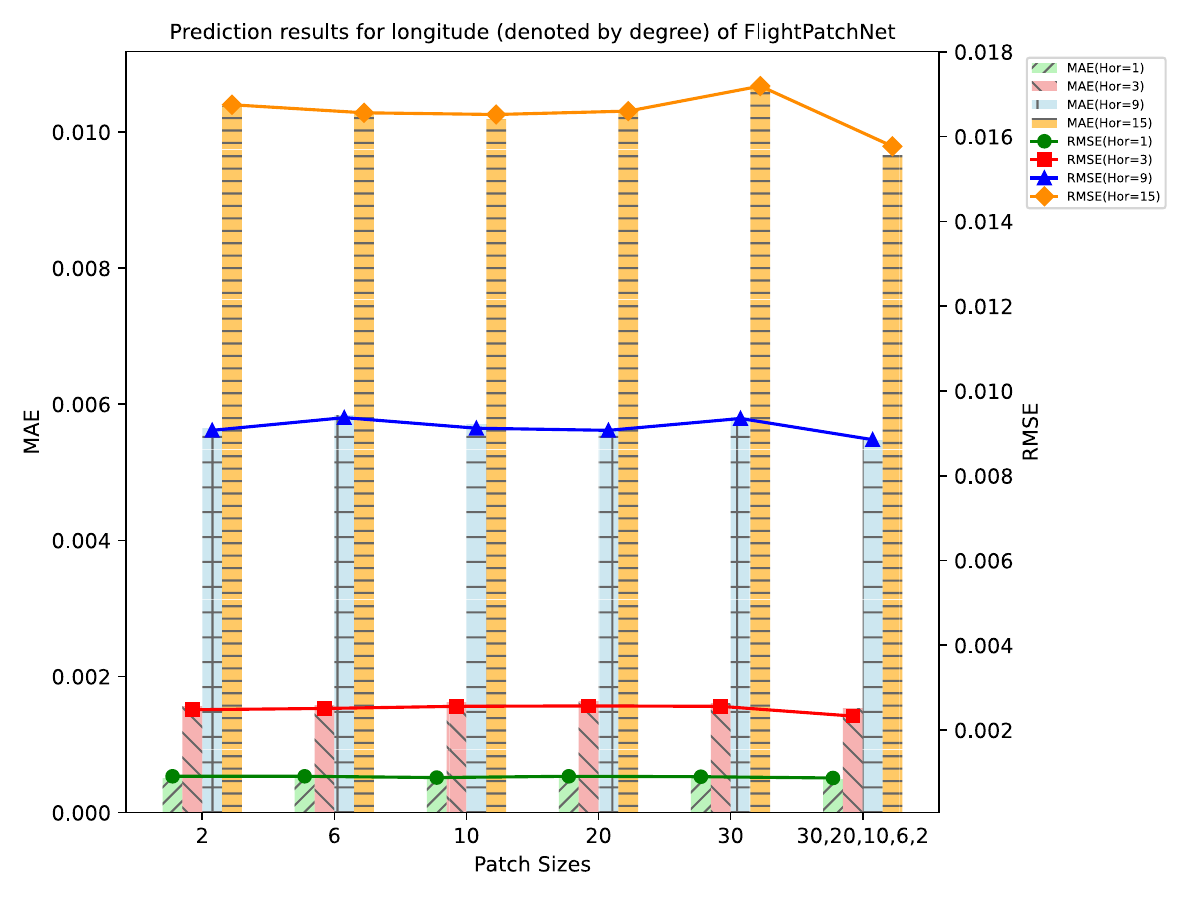}

    \captionsetup{font=small}
    \caption{MAE and RMSE of FlightPatchNet for longitude with single scale and multi scales ($L=60$).}
    \label{fig:effect_multi_scale_lon}
\end{figure}
\begin{figure}[h]
     \centering
        \includegraphics[width=0.8\linewidth]{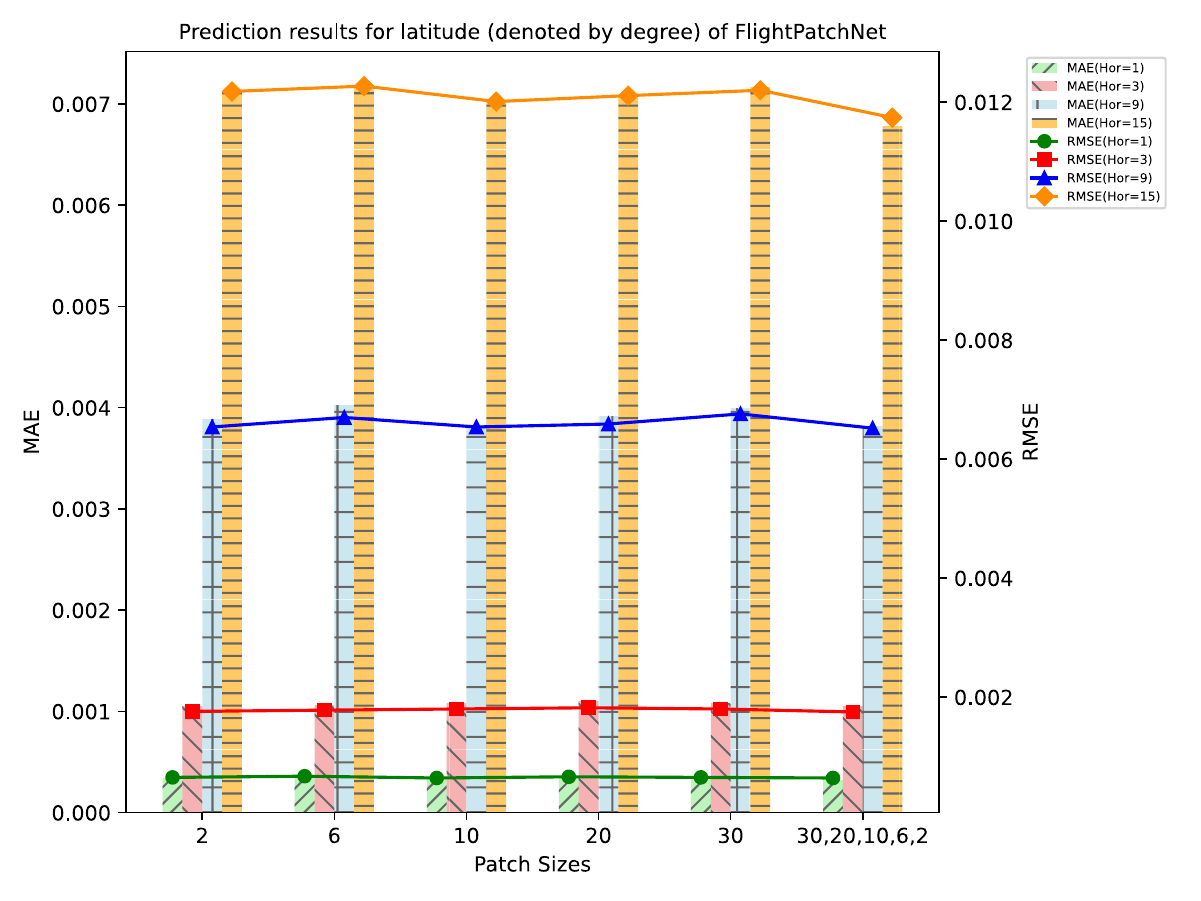}

    \captionsetup{font=small}
    \caption{MAE and RMSE of FlightPatchNet for latitude with single scale and multi scales ($L=60$).}
    \label{fig:effect_multi_scale_lat}
\end{figure}
\begin{figure}[h]
     \centering
        \includegraphics[width=0.8\linewidth]{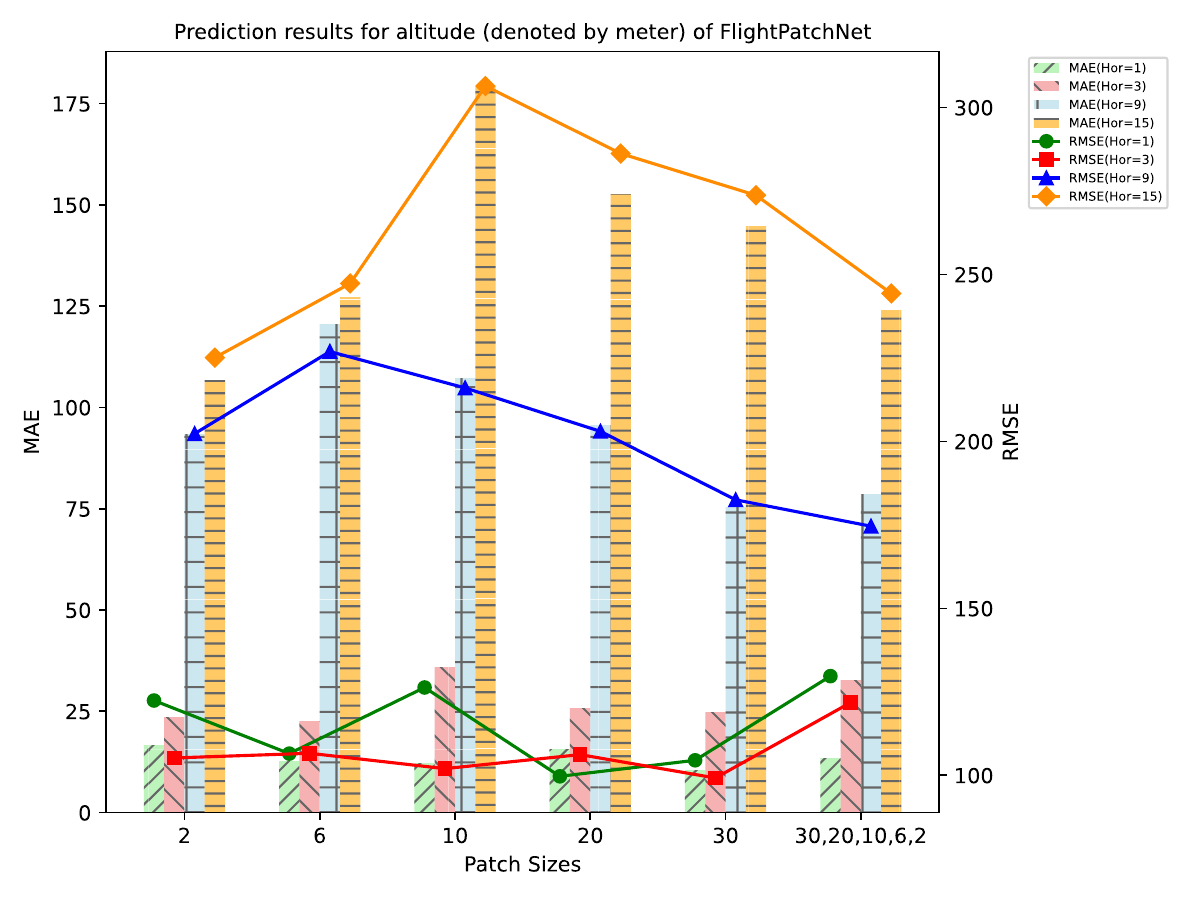}
    \captionsetup{font=small}
    \caption{MAE and RMSE of FlightPatchNet for altitude with single scale and multi scales ($L=60$).}
     \label{fig:effect_multi_scale_alt}
\end{figure}
To investigate the effect of multi-scale modeling, we conduct experiments on single scale for \{2,6,10,20,30\}. The results in Figure~\ref{fig:effect_multi_scale_lon}, \ref{fig:effect_multi_scale_lat} and \ref{fig:effect_multi_scale_alt} illustrate the critical contribution of multi scales to our model. We observe that different variables prefer distinct time scales. For instance, a patch size of 10 obtains the second-best prediction performance on longitude and latitude but the worst performance on altitude when prediction horizon $T=15$. This indicates that longitude, latitude and altitude exhibit distinct temporal patterns, and different scales can extract diverse complementary features, which can be effectively leveraged to obtain competitive and robust prediction performance. 

\subsection{Ablation Study}
We conduct ablation studies by removing corresponding modules from FlightPatchNet. Specifically, \textbf{w/o global temporal attention} does not capture the correlations between time steps. \textbf{w/o scale fusion} considers each time scale of equal importance. \textbf{w/o channel fusion} does not explore the relationships between variables. 
Table~\ref{tab:ablation_block} shows the contribution of each component.
\begin{table}[h]
\centering
\captionsetup{font=small}
\caption{Performance comparisons on ablative variants. The best results are highlighted in {\textbf{bold}}. Hor represents the prediction horizon $T\in\{1,3,9,15\}$.}
\label{tab:ablation_block}
\setlength{\tabcolsep}{1mm}{
    \scalebox{0.72}{
       \begin{tabular}{@{}c|c|cccccc@{}}
\toprule
\multirow{2}{*}{Case}                                                                         & \multirow{2}{*}{Hor} & \multicolumn{2}{c|}{\begin{tabular}[c]{@{}c@{}}Lon\\ ($0.00001^ \circ$)\end{tabular}} & \multicolumn{2}{c|}{\begin{tabular}[c]{@{}c@{}}Lat\\ ($0.00001^\circ$)\end{tabular}} & \multicolumn{2}{c}{\begin{tabular}[c]{@{}c@{}}Alt\\ (m)\end{tabular}} \\ \cmidrule(l){3-8} 
                                                                                              &                      & \multicolumn{1}{c|}{MAE}                  & \multicolumn{1}{c|}{RMSE}                 & \multicolumn{1}{c|}{MAE}                 & \multicolumn{1}{c|}{RMSE}                 & \multicolumn{1}{c|}{MAE}                & RMSE                        \\ \midrule 
\multirow{4}{*}{\begin{tabular}[c]{@{}c@{}}w/o \\ global\\ temporal\\ attention\end{tabular}} & 1                    & 51                                        & 90                                        & 34                                       & 66                                        & 21                                      & 137                         \\
                                                                                              & 3                    & 190                                       & 308                                       & 132                                      & 222                                       & \textbf{28}                             & 112                         \\
                                                                                              & 9                    & 667                                       & 1085                                      & 466                                      & 791                                       & \textbf{67}                             & \textbf{164}                \\
                                                                                              & 15                   & 1232                                      & 2005                                      & 876                                      & 1486                                      & \textbf{112}                            & \textbf{221}                \\ \midrule
\multirow{4}{*}{\begin{tabular}[c]{@{}c@{}}w/o\\ scale\\ fusion\end{tabular}}                 & 1                    & 53                                        & 92                                        & 35                                       & 67                                        & 24                                      & 130                         \\
                                                                                              & 3                    & 169                                       & 268                                       & 114                                      & 188                                       & 33                                      & \textbf{100}                \\
                                                                                              & 9                    & 609                                       & 975                                       & 409                                      & 688                                       & 91                                      & 194                         \\
                                                                                              & 15                   & 1112                                      & 1787                                      & 759                                      & 1280                                      & 162                                     & 282                         \\ \midrule
\multirow{4}{*}{\begin{tabular}[c]{@{}c@{}}w/o\\ channel\\ fusion\end{tabular}}               & 1                    & 50                                        & 89                                        & 34                                       & 65                                        & 20                                      & 160                         \\
                                                                                              & 3                    & 166                                       & 265                                       & 112                                      & 187                                       & 29                                      & 122                         \\
                                                                                              & 9                    & 573                                       & 924                                       & 398                                      & 667                                       & 73                                      & 174                         \\
                                                                                              & 15                   & 1059                                      & 1707                                      & 727                                      & 1240                                      & 132                                     & 250                         \\ \midrule
\multirow{4}{*}{FlightPatchNet}                                                               & 1                    & \textbf{48}                               & \textbf{87}                               & \textbf{32}                              & \textbf{64}                               & \textbf{13}                             & \textbf{124}                \\
                                                                                              & 3                    & \textbf{153}                              & \textbf{233}                              & \textbf{105}                             & \textbf{175}                              & 33                                      & 122                         \\
                                                                                              & 9                    & \textbf{546}                              & \textbf{885}                              & \textbf{381}                             & \textbf{652}                              & 79                                      & 175                         \\
                                                                                              & 15                   & \textbf{966}                              & \textbf{1577}                             & \textbf{678}                             & \textbf{1174}                             & 124                                     & 244                         \\ \bottomrule
\end{tabular}
    }
}
\end{table}
Removing the global temporal attention dramatically decreases the multi-step prediction performance, demonstrating the necessary of capturing the correlations between different time steps. Scale fusion can effectively improve the prediction accuracy, indicating that different time scales of trajectory series contain rich and diverse temporal variation information. Channel fusion also improves the model performance, suggesting the importance of exploring relationships between different variables in complex temporal modeling. 

\subsection{Model Complexity}
\begin{table}[h]
    \centering
    \captionsetup{font=small}
\caption{Model Complexity Comparisons. The look-back window size $L = 60$ and the prediction horizon $T = 15$ for all models.}
\label{table:modelcomplexity}
\setlength{\tabcolsep}{1mm}{
    \scalebox{0.72}{
     \begin{tabular}{@{}l|c|c|c|c@{}}
     \toprule

Type                 & Models               & \begin{tabular}[c]{@{}c@{}}Parameters\\ (MB)\end{tabular} & \begin{tabular}[c]{@{}c@{}}FLOPs\\ (M)\end{tabular} & \begin{tabular}[c]{@{}c@{}}Running Time\\ (s/iter)\end{tabular} \\ \midrule 
\multirow{2}{*}{DMS} & FlightPatchNet & 5.69                                                      & 64.38                                               & 0.0069                                                          \\ \cmidrule(l){2-5} 
                     & FlightBERT++         & 44.26                                                     & 3000.00                                                & 0.0112 \\ \cmidrule(l){2-5} 
                     & Pathformer         & 2.64                                                     & 258.55                                               & 0.02174  \\ \cmidrule(l){2-5} 
                     & TimeMixer         & 0.45                                                    & 3115.00                                                & 0.0037    \\ \cmidrule(l){2-5} 
                     & TimesNet         & 37.50                                                    & 196159.87                                              & 0.0534    \\   \cmidrule(l){2-5}                            
                     & MICN         & 1.28                                                    & 1235.83                                              & 0.0012
                     \\ \midrule
\multirow{4}{*}{IMS}  &FlightBERT           & 25.31                                                     & 1620.00                                                & 0.2406                                                          \\ \cmidrule(l){2-5} 
                      & LSTM                 & 0.03                                                      & 1.67                                                & 0.0583                                                          \\ \cmidrule(l){2-5} 
                     & Bi-LSTM              & 0.51                                                      & 31.15                                               & 0.1241                                                          \\ \cmidrule(l){2-5} 
                     & CNN-LSTM             & 0.04                                                      & 1.22                                                & 0.0429  
                      \\ \cmidrule(l){2-5} 
                     & WTFTP            & 0.23                                                     & 60.00                                               & 0.0290\\ \bottomrule 
    \end{tabular}
    }
}
\end{table}

As shown in Table~\ref{table:modelcomplexity}, our proposed FlightPatchNet achieves the greatest efficiency and has relatively fewer parameters compared to other models. For multi-step prediction, the DMS-based models demonstrate significant improvements in computational performance compared to the IMS-based models. In addition, FlightPatchNet is lightweight compared to FlightBERT++ and FlightBERT, which indicates our model can provide a promising solution for real-time air transportation management. 

\section{Conclusion, Limitation and Future Work}
\paragraph{Conclusion} In this paper, we propose FlightPatchNet, a multi-scale patch network with differential coding for the short-term FTP task. The differential coding is leveraged to reduce the significant differences in the original data range and reflect the temporal variations in realistic flight trajectories. The multi-scale patch network is designed to explore global trends and local details based on divided patches of different sizes, and integrate scale-wise correlations and inter-variable relationships for complete temporal modeling. Extensive experiments on a real-world dataset demonstrate that FlightPatchNet achieves the most competitive performance and offers a significant reduction in computational complexity, presenting a promising solution for real-time air traffic control applications.
\paragraph{Limitation and Future Work} The original series of altitude contains many unexpected noises. Our primary focus on modeling temporal variations enables FlightPatchNet to concentrate more on these irregular change points during altitude prediction, leading to a large bias. In the future, we will further explore temporal modeling of altitude and investigate robust noise-handling techniques such as moving average to make the series smoother and less sensitive to noise. In addition, the data-missing problem occurs commonly in realistic flight trajectory series, making  downstream analysis difficult. Thus, we attempt to incorporate data imputation methods into our model to enhance the model applicability and provide a general framework for flight trajectory prediction.
\bibliography{main}
\newpage
\onecolumn

\appendix
\section*{Appendix}
\section{Experimental Details}\label{app:exp}
\subsection{Dataset Preprocessing and Description}\label{dataset}
This paper exploits real-world datasets provided by OpenSky from 2020 to 2022 to validate our proposed model. The data preprocessing steps are as follows:

(1) Data Extraction: We extract seven features from the raw data, including timestamp, longitude, latitude, altitude, horizontal flight speed, horizontal flight angle, and vertical speed. The timestamp is used to identify whether the trajectory points are continuous, and the other six features are further processed as inputs to the model.

(2) Data Filtering: Due to many missing values and outliers in the raw dataset, we select 100 consecutive points without missing values as a complete flight trajectory. Then, we adopt the z-score method to find out the outliers. If one flight trajectory contains any outliers, we discard the whole trajectory. The z-score formula is as follows:
\begin{equation}
	z = \frac{(\overline{x}-\mu)}{\sigma - \sqrt{n}}
	\tag{1}
\end{equation}
where $\overline{x}$ is the value of each feature point, $\mu$ is the mean of each feature, $\sigma$ is the variance of each feature, and $n$ is the number of feature points.

(3) Velocity Transformation: We transform the horizontal velocity into $V_x$ and $V_y$ according to the angle, where $V_x$ is the velocity in the longitude dimension and $V_y$ is the velocity in the latitude dimension. In this way, the features become longitude, latitude, altitude, $V_x$, $V_y$ and $V_z$.

(4) Data Segmentation: The dataset is randomly divided into three parts with a ratio of  8:1:1 for training, validation, and testing.

After the above preprocessing, 274,605 flight trajectories are selected into our dataset. The range of longitude, latitude and altitude are $[-179.86396^\circ, 178.82147^\circ]$, $[-46.42435^\circ, 70.32590^\circ]$ and $[0, 21031.00m]$, respectively. The interval between two adjacent flight trajectory points is 10 seconds.

\subsection{Baseline Methods}\label{baseline}
We briefly describe the selected 10 competitive baselines as follows:
\begin{itemize}
    \item LSTM \citep{LSTM8489734}: Based on two layers of LSTM (with 30 and 60 nodes respectively) to encode each trajectory point, and future trajectories are predicted through a fully connected layer.
    \item CNN-LSTM \citep{CNN-LSTM9145522}: Based on two layers of one-dimensional CNN (the convolution kernel size is $1 \times 3$) and two layers of LSTM (with 50 nodes) to encode each trajectory point, and future trajectories are predicted through a fully connected layer.
    \item Bi-LSTM \citep{Sahadevan}: Based on two layers of Bi-LSTM (with 200 and 50 nodes respectively) to encode each trajectory point, and future trajectories are predicted through a fully connected layer.
    \item FlightBERT \citep{Guo2023FlightBERT}: It utilizes a BE representation to convert the scalar attributes of the flight trajectory into binary vectors, considering the FTP task as a multi binary classification problem. It uses 18, 16, 11 and 11 bits to encode the real values (decimals) of longitude, latitude, altitude and velocities into BE representation respectively. 
    \item FlightBERT++ \citep{Guo2023FlightBERT++}: It inherits the BE representation from the FlightBERT and introduces a differential prediction paradigm, which aims to predict the differential values of the trajectory attributes instead of the absolute values.
    \item WTFTP \citep{Zhang2023FlightTP}: It is an IMS-based method that utilizes discrete wavelet transform (DWT) to decompose the input flight trajectory into wavelet coefficients and predicts future trajectories based on the generated wavelet coefficients by inverse discrete wavelet transform (IDWT).
    \item TimeMixer \citep{wang2024timemixer}: It is a fully MLP-based architecture that mixes the decomposed seasonal and trend components in fine-to-coarse and coarse-to-fine directions separately and ensembles multiple predictors to utilize complementary forecasting capabilities in multi-scale observations.
    \item TimesNet \citep{wu2022timesnet}: It is a task-general foundational model for time series analysis, which disentangles complex temporal variations into multiple intra-period and inter-period variations. A parameter efficient inception block is employed to capture these temporal variations in 2D space.
    \item MICN \citep{wang2023micn}: It adopts a multi-scale branch structure to capture the underlying information in time series. Downsampling one-dimensional convolution is used for local feature extraction and isometric convolution is employed for global correlation discovery.
    \item Pathformer \citep{chen2024pathformer}: It is a multi-scale Transformer with adaptive pathways, which integrates both temporal resolution and temporal distance for multi-scale modeling. A multi-scale router with temporal decomposition and an aggregator work together to realize adaptive multi-scale modeling for time series.   
\end{itemize}

\subsection{Implementation Details}\label{app:implementation}
For fairness, all the models follow the same experimental setup with look-back window $L = 60$ and prediction horizon $T\in\{1,3,9,15\}$,  which means the observation time is 10 minutes and the forecasting time is 10 seconds, 30 seconds, 1.5 minutes, 2.5 minutes. The patch sizes in multi-scale patch mixer blocks are set to \{30, 20, 10, 6, 2\}. The dimension of temporal embedding $d$ is 128. For all the MSA in this paper, the head number is 8 and the attention layer $l$ is 3. The learning rate is set as $10^{-4}$ for all experiments.  Our method is trained with MSE loss, using the Adam optimizer. The training process is early stopped within 30 epochs. The training would be terminated early if the validation loss does not decrease for three consecutive rounds. The model is implemented in PyTorch 2.2.1 and trained on a single NVIDIA RTX 3090 GPU with 24GB memory.

\subsection{Evaluation Metrics}\label{metrics}
Mean Absolute Error (MAE) and Root Mean Squared Error (RMSE) are exploited to evaluate the proposed model and baselines, which are defined as:
\begin{equation*}
    \begin{aligned} 
    MAE &=\frac{1}{T}\sum_{i=1}^T|\mathbf{Y}_i-\hat{\mathbf{Y}}_i| \\
    RMSE&=\sqrt{\frac{1}{T}{\sum_{i=1}^T(\mathbf{Y}_i-\hat{\mathbf{Y}}_i)^2}}
    \end{aligned}
\end{equation*}
where  $\mathbf{Y}_i$, $\hat{\mathbf{Y}}_i$  are the ground truth and prediction result for  $i$-th future point, respectively.

\section{Additional Experimental Results}
\subsection{Hyper-Parameter Sensitivity }
\begin{figure}[htbp]
    \centering
    \includegraphics[width=0.8\linewidth]{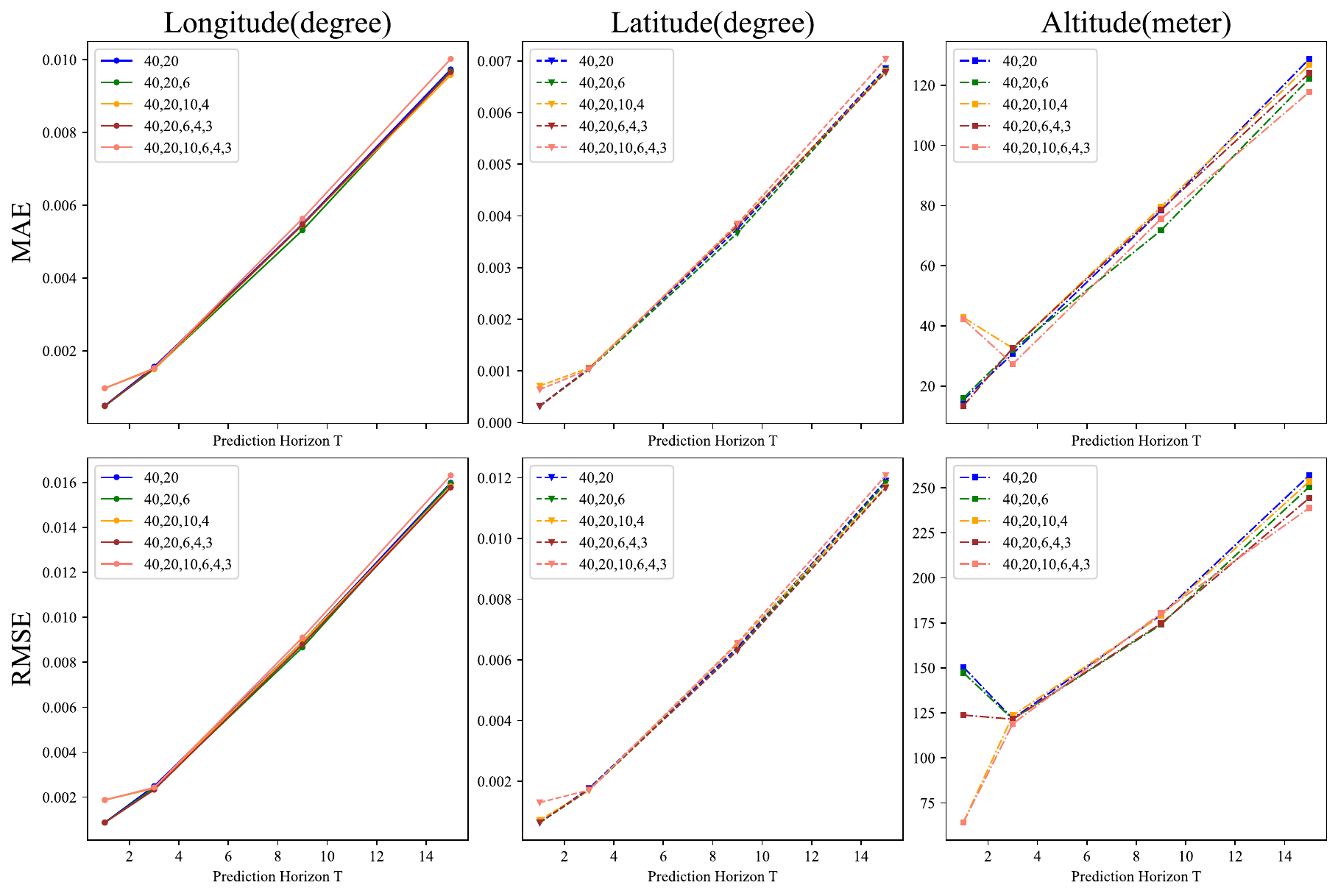}
   \captionsetup{font=small}
    \caption{MAE and RMSE with different number of scales for prediction horizon $T \in \{1,3,9,15\}$.}
    \label{fig:different_scales_rmse}
\end{figure}

\paragraph{Number of Scales}
We perform experiments on different number of scales and report the MAE and RMSE results. As shown in Figure~\ref{fig:different_scales_rmse}, we can observe that when the number of scales increases from 2 to 5, the performance of FlightPatchNet is constantly improved. This is because FlightPatchNet can capture diverse global and local temporal patterns under different scales. When the number of scales increases up to 6, the performance starts to deteriorate. This indicates that a certain number of scales is sufficient for temporal modeling, and excessive scales may lead to the overfitting problem.

\paragraph{Number of Attention Layers }

\begin{figure}[htbp]
    \centering
    \subfigure[MAE and RMSE of different attention layers in global temporal attention.]{
       \centering
       \includegraphics[width=0.3\linewidth]{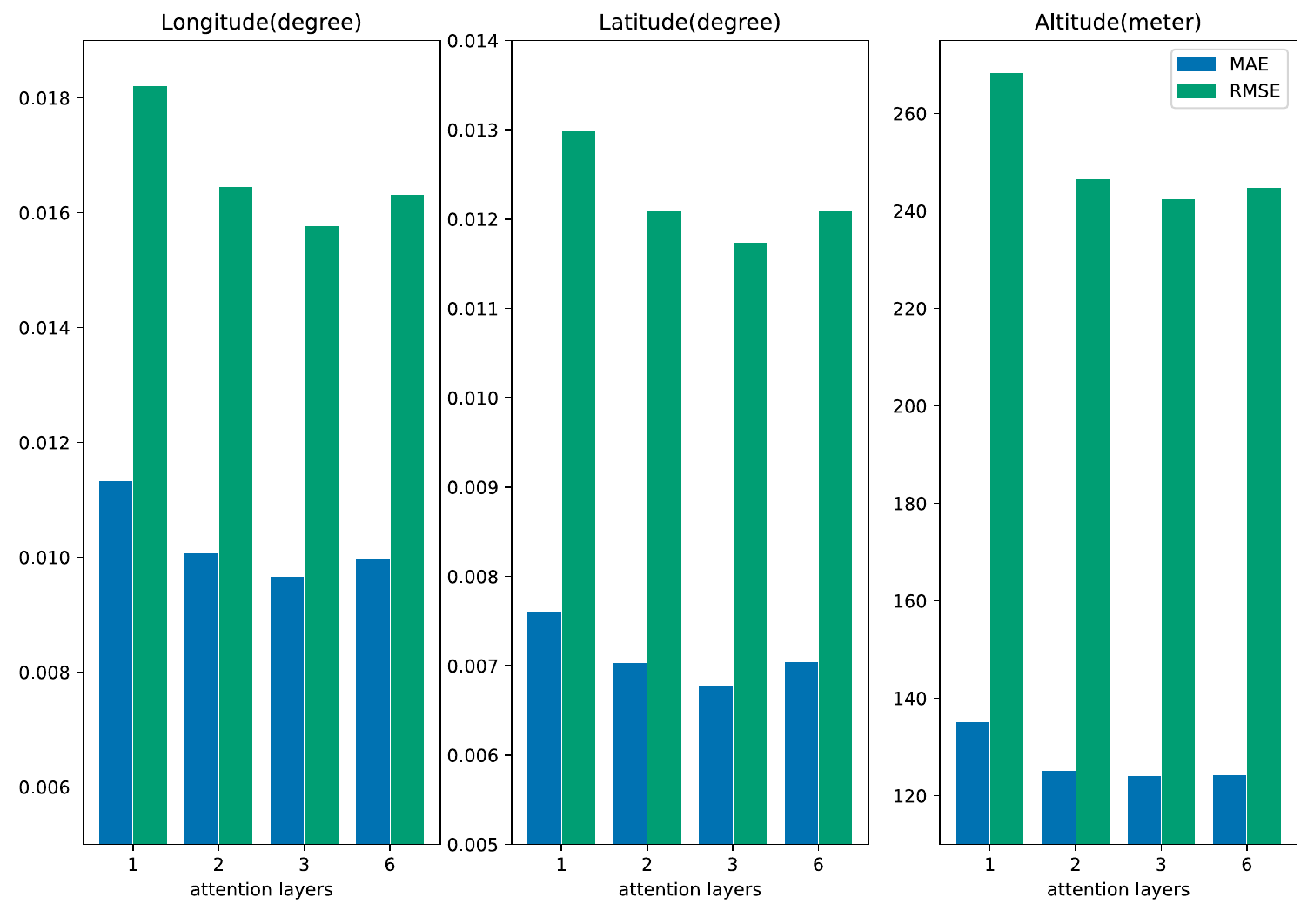}  
        \label{fig:different_temporal_attention_layer} 
    }
    \subfigure[MAE and RMSE of different attention layers in scale fusion.]{
     \centering
    \includegraphics[width=0.3\linewidth]{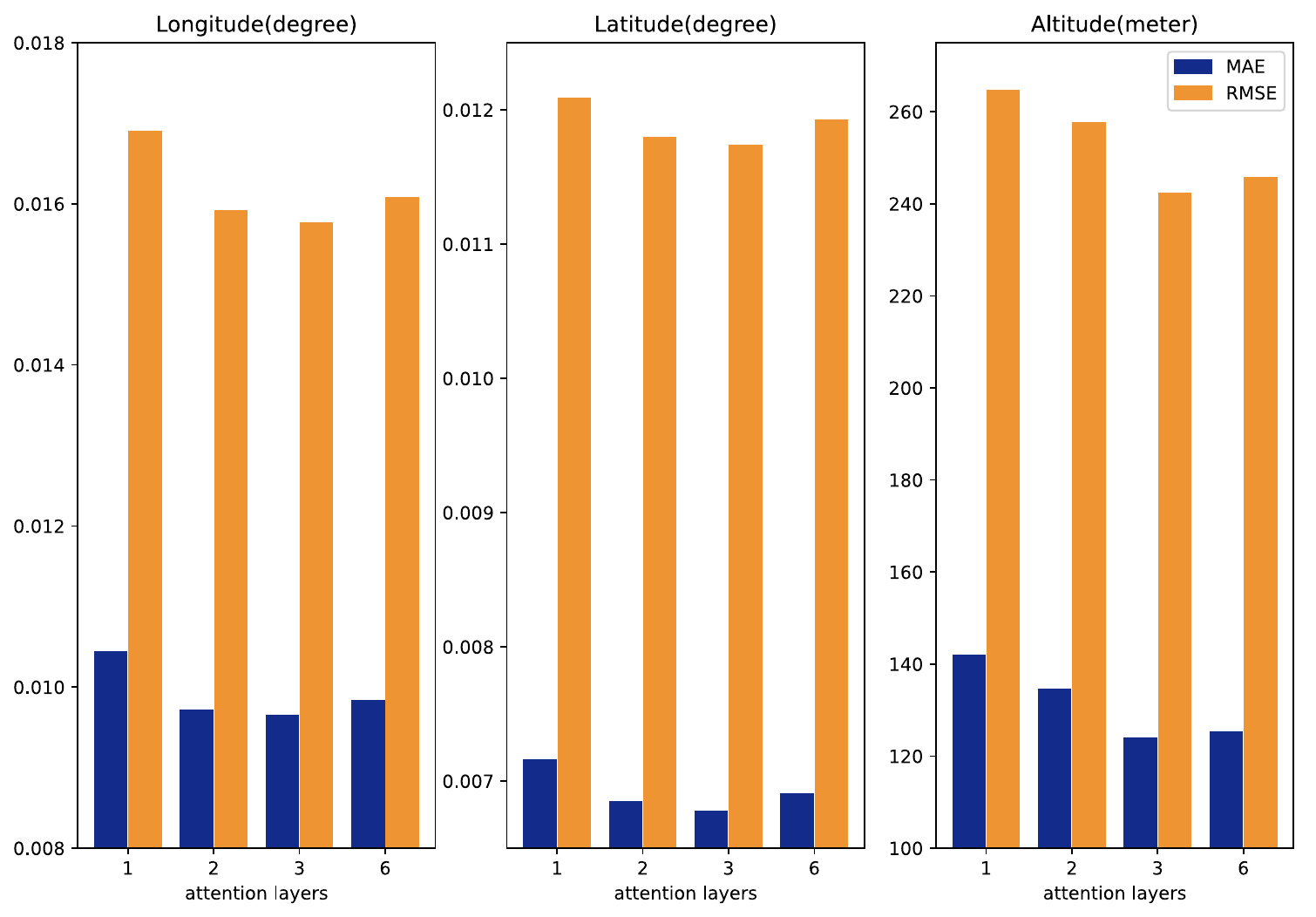} 
        \label{fig:different_scale_attention_layer} 
    }
    \subfigure[MAE and RMSE of different attention layers in channel fusion.]{
     \centering
    \includegraphics[width=0.3\linewidth]{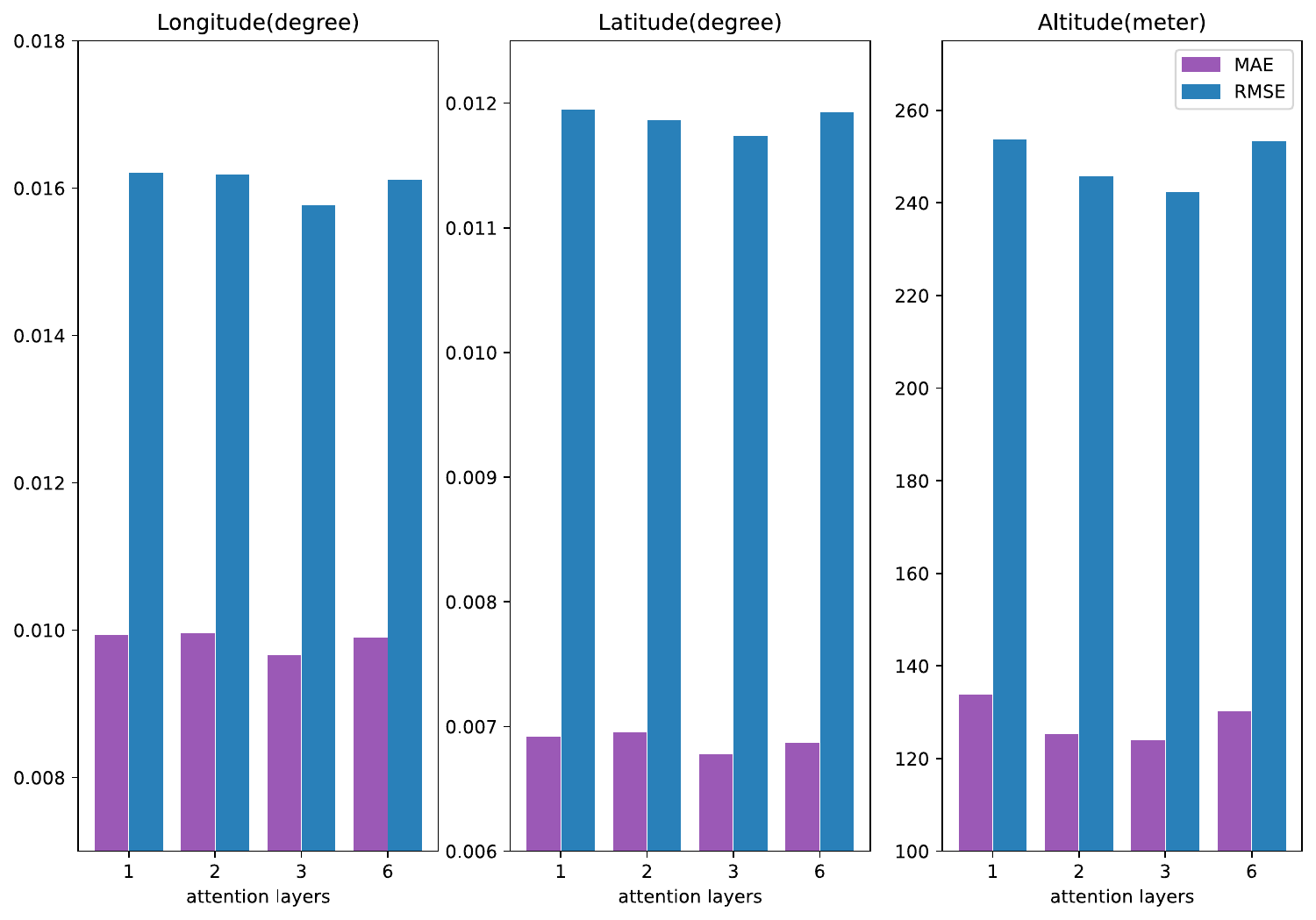}  
        \label{fig:different_channel_attention_layer} 
    }
    \captionsetup{font=small}
    \caption{MAE and RMSE of different attention layers for prediction horizon $T=15$}
    \label{fig:Attention_layers}
 \end{figure}

We test the number of attention layers in $\{1,2,3,6\}$ for global temporal attention, scale fusion, and channel fusion. The results are shown in Figure~\ref{fig:different_temporal_attention_layer}, Figure~\ref{fig:different_scale_attention_layer}  and Figure~\ref{fig:different_channel_attention_layer}. We can observe that when the number of attention layers increases from 1 to 3, the values of MAE and RMSE decrease, demonstrating that our model can better capture the dependencies between different time steps, scale-wise correlations and inter-variable relationships with more layers of attention. When the number of attention layers increases up to 6, the prediction accuracy does not improve. Thus, we choose to use three layers of attention in these parts. 
 
\begin{figure}[htbp]
    \centering
    \subfigure[]{
        \centering
       \includegraphics[width=0.45\linewidth]{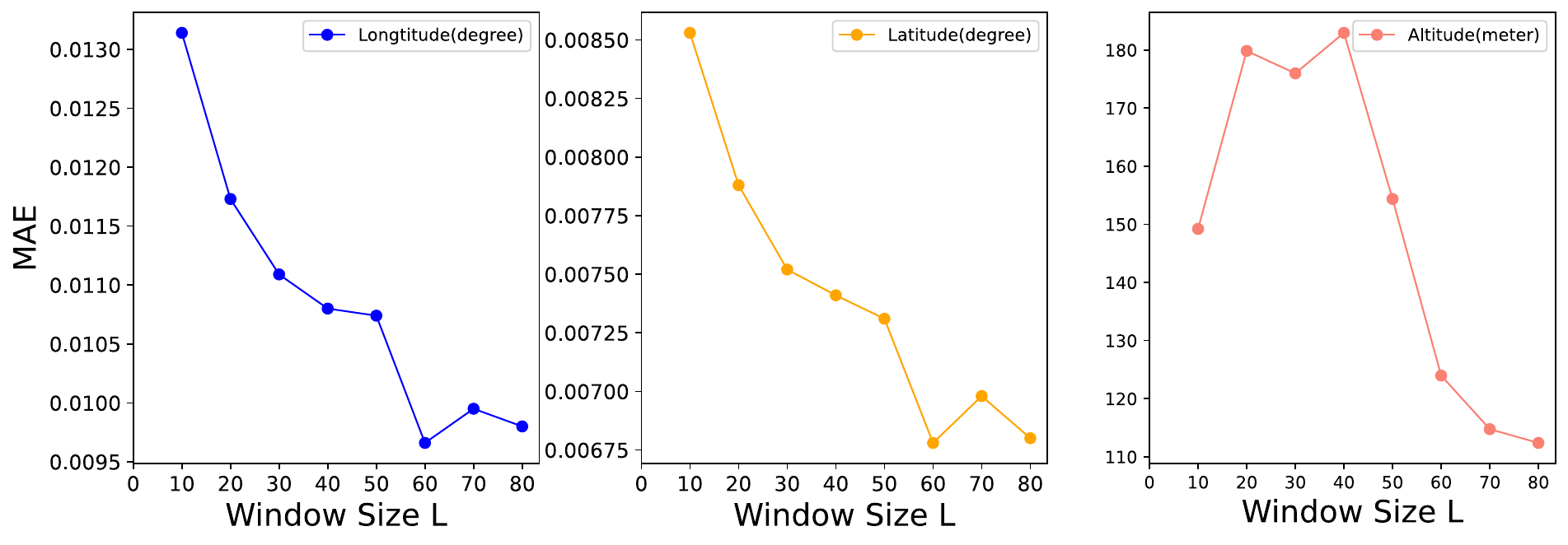}    
    }
    \subfigure[]{
        \centering
        \includegraphics[width=0.45\linewidth]{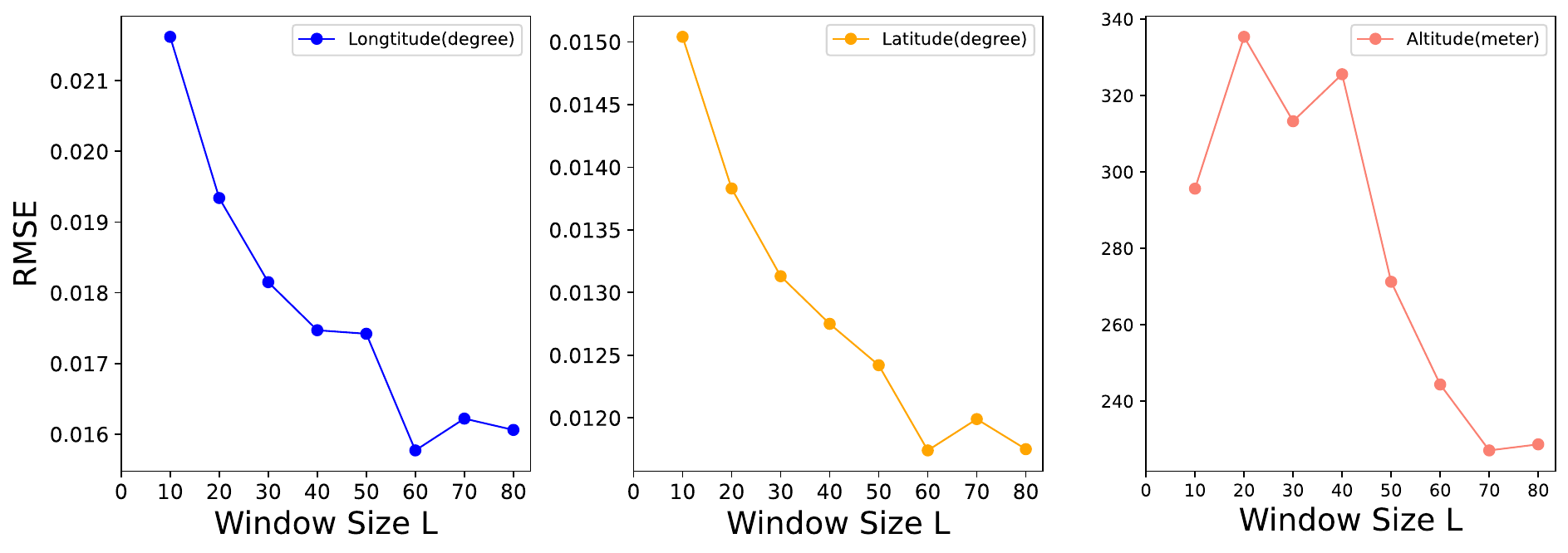}    
    }
    \captionsetup{font=small}
    \caption{MAE and RMSE of different look-back window sizes $L$ for prediction horizon $T=15$.}
    \label{fig:different_input_length}
\end{figure}

\paragraph{Look-Back Window Size $\mathbf{L}$}
Figure~\ref{fig:different_input_length} demonstrates the MAE and RMSE results of our model with different look-back window sizes. We set the window size $L$ to $\{10, 20, 30, 40, 50, 60, 70, 80\}$. The overall performance of FlightPatchNet is significantly improved as the window size increases from 10 to 60, indicating that FlightPatchNet can thoroughly capture the temporal dependencies from long flight trajectories. Moreover, the performance of altitude fluctuates with the increase of the window size, suggesting that the series of altitude are non-stationary and easily affected by unexpected noise. Thus, we set $L$ as 60 to achieve the overall optimal performance.

\paragraph{Order of Scales}
We conduct experiments on the order of patch sizes and report the MAE and RMSE results. As shown in Table~\ref{table:effective of inverted pyramid}, we can observe that patch sizes in descending order can effectively improve the prediction performance, indicating that the macro knowledge from coarser scales can guide the temporal modeling of finer scales.
\begin{table*}[ht]
    \centering
    \captionsetup{font=small}
    \caption{ The results of flight trajectory prediction with scales in ascending and descending order. $\uparrow$ means scales in ascending order and $\downarrow$ means scales in descending order.  The better results are highlighted in \textbf{bold}.}
    \label{table:effective of inverted pyramid}
    \setlength{\tabcolsep}{1mm}{
        {
            \begin{tabular}{@{}c|cccccc|cccc|cccc@{}}
    \toprule
    \multirow{2}{*}{patch sizes}  & \multicolumn{6}{c|}{Lon($0.00001^\circ$)}                                                                                                                                                                     & \multicolumn{4}{c|}{Lat($0.00001^\circ$)}                                                                                 & \multicolumn{4}{c}{Alt(m)}                                                                                                           \\ \cmidrule(l){2-15} 
                                  & \multicolumn{1}{c|}{Style}                         & \multicolumn{1}{c|}{Horizon} & \multicolumn{1}{c|}{1}            & \multicolumn{1}{c|}{3}            & \multicolumn{1}{c|}{9}            & 15            & \multicolumn{1}{c|}{1}            & \multicolumn{1}{c|}{3}            & \multicolumn{1}{c|}{9}            & 15            & \multicolumn{1}{c|}{1}               & \multicolumn{1}{c|}{3}               & \multicolumn{1}{c|}{9}               & 15              \\ \midrule
    \multirow{4}{*}{2,6,10,20,30} & \multicolumn{1}{c|}{\multirow{2}{*}{$\uparrow$}}   & \multicolumn{1}{c|}{MAE}     & \multicolumn{1}{c|}{98}           & \multicolumn{1}{c|}{155}          & 548                               & 1008          & \multicolumn{1}{c|}{99}           & \multicolumn{1}{c|}{106}          & \multicolumn{1}{c|}{385}          & 697           & \multicolumn{1}{c|}{54.02}           & \multicolumn{1}{c|}{33.47}           & \multicolumn{1}{c|}{79.39}           & 127.51          \\ \cmidrule(l){3-15} 
                                  & \multicolumn{1}{c|}{}                              & \multicolumn{1}{c|}{RMSE}    & \multicolumn{1}{c|}{187}          & \multicolumn{1}{c|}{241}          & \multicolumn{1}{c|}{887}          & 1642          & \multicolumn{1}{c|}{131}          & \multicolumn{1}{c|}{183}          & \multicolumn{1}{c|}{656}          & 1197          & \multicolumn{1}{c|}{\textbf{81.92}}  & \multicolumn{1}{c|}{\textbf{110.68}} & \multicolumn{1}{c|}{184.54}          & 248.16          \\ \cmidrule(l){2-15} 
                                  & \multicolumn{1}{c|}{\multirow{2}{*}{$\downarrow$}} & \multicolumn{1}{c|}{MAE}     & \multicolumn{1}{c|}{\textbf{48}}  & \multicolumn{1}{c|}{\textbf{153}} & \multicolumn{1}{c|}{\textbf{546}} & \textbf{966}  & \multicolumn{1}{c|}{\textbf{32}}  & \multicolumn{1}{c|}{\textbf{105}} & \multicolumn{1}{c|}{\textbf{381}} & \textbf{678}  & \multicolumn{1}{c|}{\textbf{13.34}}  & \multicolumn{1}{c|}{\textbf{32.65}}  & \multicolumn{1}{c|}{\textbf{78.57}}  & \textbf{123.97} \\ \cmidrule(l){3-15} 
                                  & \multicolumn{1}{c|}{}                              & \multicolumn{1}{c|}{RMSE}    & \multicolumn{1}{c|}{\textbf{87}}  & \multicolumn{1}{c|}{\textbf{233}} & \multicolumn{1}{c|}{\textbf{885}} & \textbf{1577} & \multicolumn{1}{c|}{\textbf{64}}  & \multicolumn{1}{c|}{\textbf{175}} & \multicolumn{1}{c|}{\textbf{652}} & \textbf{1174} & \multicolumn{1}{c|}{129.65}          & \multicolumn{1}{c|}{121.78}          & \multicolumn{1}{c|}{\textbf{174.63}} & \textbf{244.34} \\ \midrule
    \multirow{4}{*}{3,4,6,20,40}  & \multicolumn{1}{c|}{\multirow{2}{*}{$\uparrow$}}   & \multicolumn{1}{c|}{MAE}     & \multicolumn{1}{c|}{98}           & \multicolumn{1}{c|}{155}          & \multicolumn{1}{c|}{556}          & 997           & \multicolumn{1}{c|}{64}           & \multicolumn{1}{c|}{105}          & \multicolumn{1}{c|}{383}          & 704           & \multicolumn{1}{c|}{\textbf{39.77}}  & \multicolumn{1}{c|}{32.15}           & \multicolumn{1}{c|}{\textbf{76.38}}  & \textbf{124.18} \\ \cmidrule(l){3-15} 
                                  & \multicolumn{1}{c|}{}                              & \multicolumn{1}{c|}{RMSE}    & \multicolumn{1}{c|}{188}          & \multicolumn{1}{c|}{247}          & \multicolumn{1}{c|}{901}          & 1631          & \multicolumn{1}{c|}{131}          & \multicolumn{1}{c|}{175}          & \multicolumn{1}{c|}{655}          & 1210          & \multicolumn{1}{c|}{64.86}           & \multicolumn{1}{c|}{124.20}          & \multicolumn{1}{c|}{177.46}          & \textbf{243.46} \\ \cmidrule(l){2-15} 
                                  & \multicolumn{1}{c|}{\multirow{2}{*}{$\downarrow$}} & \multicolumn{1}{c|}{MAE}     & \multicolumn{1}{c|}{\textbf{97}}  & \multicolumn{1}{c|}{\textbf{153}} & \multicolumn{1}{c|}{\textbf{542}} & \textbf{963}  & \multicolumn{1}{c|}{\textbf{63}}  & \multicolumn{1}{c|}{\textbf{104}} & \multicolumn{1}{c|}{\textbf{369}} & \textbf{670}  & \multicolumn{1}{c|}{43.46}           & \multicolumn{1}{c|}{\textbf{28.96}}  & \multicolumn{1}{c|}{79.27}           & 128.13          \\ \cmidrule(l){3-15} 
                                  & \multicolumn{1}{c|}{}                              & \multicolumn{1}{c|}{RMSE}    & \multicolumn{1}{c|}{\textbf{187}} & \multicolumn{1}{c|}{\textbf{245}} & \multicolumn{1}{c|}{\textbf{879}} & \textbf{1582} & \multicolumn{1}{c|}{\textbf{130}} & \multicolumn{1}{c|}{\textbf{174}} & \multicolumn{1}{c|}{\textbf{631}} & \textbf{1167} & \multicolumn{1}{c|}{\textbf{64.50}}  & \multicolumn{1}{c|}{\textbf{115.76}} & \multicolumn{1}{c|}{\textbf{176.96}} & 251.72          \\ \midrule
    \multirow{4}{*}{3,6,40}       & \multicolumn{1}{c|}{\multirow{2}{*}{$\uparrow$}}   & \multicolumn{1}{c|}{MAE}     & \multicolumn{1}{c|}{48}           & \multicolumn{1}{c|}{156}          & \multicolumn{1}{c|}{536}          & 994           & \multicolumn{1}{c|}{35}           & \multicolumn{1}{c|}{105}          & \multicolumn{1}{c|}{370}          & 691           & \multicolumn{1}{c|}{\textbf{14.96}}  & \multicolumn{1}{c|}{31.42}           & \multicolumn{1}{c|}{79.07}           & \textbf{117.43} \\ \cmidrule(l){3-15} 
                                  & \multicolumn{1}{c|}{}                              & \multicolumn{1}{c|}{RMSE}    & \multicolumn{1}{c|}{87}           & \multicolumn{1}{c|}{248}          & \multicolumn{1}{c|}{876}          & 1628          & \multicolumn{1}{c|}{65}           & \multicolumn{1}{c|}{176}          & \multicolumn{1}{c|}{634}          & 1193          & \multicolumn{1}{c|}{\textbf{107.43}} & \multicolumn{1}{c|}{118.36}          & \multicolumn{1}{c|}{177.40}          & 238.87          \\ \cmidrule(l){2-15} 
                                  & \multicolumn{1}{c|}{\multirow{2}{*}{$\downarrow$}} & \multicolumn{1}{c|}{MAE}     & \multicolumn{1}{c|}{\textbf{48}}  & \multicolumn{1}{c|}{\textbf{153}} & \multicolumn{1}{c|}{\textbf{534}} & \textbf{988}  & \multicolumn{1}{c|}{\textbf{33}}  & \multicolumn{1}{c|}{\textbf{103}} & \multicolumn{1}{c|}{\textbf{368}} & \textbf{685}  & \multicolumn{1}{c|}{16.81}           & \multicolumn{1}{c|}{\textbf{31.26}}  & \multicolumn{1}{c|}{\textbf{71.96}}  & 118.66          \\ \cmidrule(l){3-15} 
                                  & \multicolumn{1}{c|}{}                              & \multicolumn{1}{c|}{RMSE}    & \multicolumn{1}{c|}{\textbf{87}}  & \multicolumn{1}{c|}{\textbf{244}} & \multicolumn{1}{c|}{\textbf{870}} & \textbf{1620} & \multicolumn{1}{c|}{\textbf{64}}  & \multicolumn{1}{c|}{\textbf{173}} & \multicolumn{1}{c|}{\textbf{633}} & \textbf{1186} & \multicolumn{1}{c|}{145.25}          & \multicolumn{1}{c|}{\textbf{114.33}} & \multicolumn{1}{c|}{\textbf{175.63}} & \textbf{236.32} \\ \bottomrule
    \end{tabular}
        }
    }
\end{table*}

\subsection{Error Bar}\label{error_bar}
In this paper, we repeat all the experiments five times. Here we report the standard deviation of our model and the second best model in Table~\ref{table:exp_res_error}. 
\begin{table*}[htbp]
\centering
\captionsetup{font=small}
\caption{Error bar of our FlightPatchNet and the second best model FlightBERT++.}
\label{table:exp_res_error}

{
\setlength{\tabcolsep}{1mm}{
 {
    \begin{tabular}{@{}c|c|cc|cc|cc@{}}
\toprule
   
\multirow{2}{*}{Model}                                                           & \multirow{2}{*}{Horizon} & \multicolumn{2}{c|}{Lon($0.00001^\circ$)}                                                         & \multicolumn{2}{c|}{Lat($0.00001^\circ$)}                                                           & \multicolumn{2}{c}{Alt(m)}                                                   \\ \cmidrule(l){3-8} 
                                                                                 &                          & \multicolumn{1}{c|}{MAE}                          & RMSE                         & \multicolumn{1}{c|}{MAE}                           & RMSE                          & \multicolumn{1}{c|}{MAE}                      & RMSE                      \\ \midrule
\multirow{4}{*}{FlightBERT++}                                                    & 1                        & \multicolumn{1}{c|}{173$\pm$6.45}          & 360$\pm$8.28& \multicolumn{1}{c|}{85$\pm$3.33}              & 148$\pm$13.20& \multicolumn{1}{c|}{9.39$\pm$1.79}   & 175.29$\pm$29.09          \\ \cmidrule(l){2-8} 
                                                                                 & 3                        & \multicolumn{1}{c|}{317$\pm$26.10}          & 659$\pm$4.35& \multicolumn{1}{c|}{210$\pm$30.50}              & 425$\pm$12.40& \multicolumn{1}{c|}{21.89$\pm$5.58}  & 167.16$\pm$46.39          \\ \cmidrule(l){2-8} 
                                                                                 & 9                        & \multicolumn{1}{c|}{871$\pm$17.40}          & 1846$\pm$44.50& \multicolumn{1}{c|}{612$\pm$60.70}              & 959$\pm$21.90& \multicolumn{1}{c|}{47.84$\pm$2.87}  & 327.93$\pm$52.84          \\ \cmidrule(l){2-8} 
                                                                                 & 15                       & \multicolumn{1}{c|}{1187$\pm$5.91}          & 3131$\pm$53.30& \multicolumn{1}{c|}{1048$\pm$36.90}           & 2127$\pm$20.10& \multicolumn{1}{c|}{78.46$\pm$8.13}  & 384.18$\pm$51.82          \\ \midrule
\multirow{4}{*}{\begin{tabular}[c]{@{}c@{}}FlightPatchNet\\ (Ours)\end{tabular}} & 1                        & \multicolumn{1}{c|}{48$\pm$1.24} & 87$\pm$1.02& \multicolumn{1}{c|}{32$\pm$1.06} & 64$\pm$0.84& \multicolumn{1}{c|}{13.34$\pm$9.43}  & 123.78$\pm$15.13\\ \cmidrule(l){2-8} 
                                                                                 & 3                        & \multicolumn{1}{c|}{153$\pm$3.19} & 233$\pm$5.44e-4& \multicolumn{1}{c|}{105$\pm$1.19}  & 175$\pm$2.36& \multicolumn{1}{c|}{32.65$\pm$1.76}  & 121.48$\pm$2.81  \\ \cmidrule(l){2-8} 
                                                                                 & 9                        & \multicolumn{1}{c|}{546$\pm$15.40} & 885$\pm$21.60& \multicolumn{1}{c|}{381$\pm$7.47}  & 652$\pm$7.76& \multicolumn{1}{c|}{78.57$\pm$2.66}  & 174.63$\pm$6.87  \\ \cmidrule(l){2-8} 
                                                                                 & 15                       & \multicolumn{1}{c|}{966$\pm$36.50} & 1577$\pm$54.90& \multicolumn{1}{c|}{678$\pm$25.80}  & 1174$\pm$35.30& \multicolumn{1}{c|}{123.97$\pm$5.72} & 244.34$\pm$6.91  \\ \bottomrule
                                                                                 
\end{tabular} 
  }
}
}
\end{table*}

\section{3D Trajectory Visualization}\label{Visualization}

\begin{figure*}[htbp]
    \subfigure[]{
    
     \includegraphics[width=0.5\linewidth]{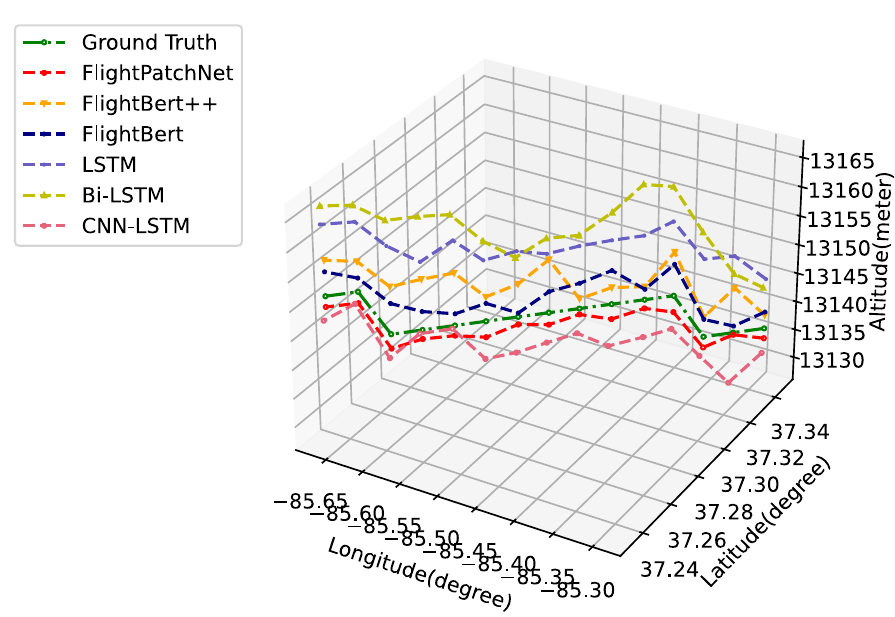}
    
    }
    \subfigure[]{
      \centering
     \includegraphics[width=0.5\linewidth]{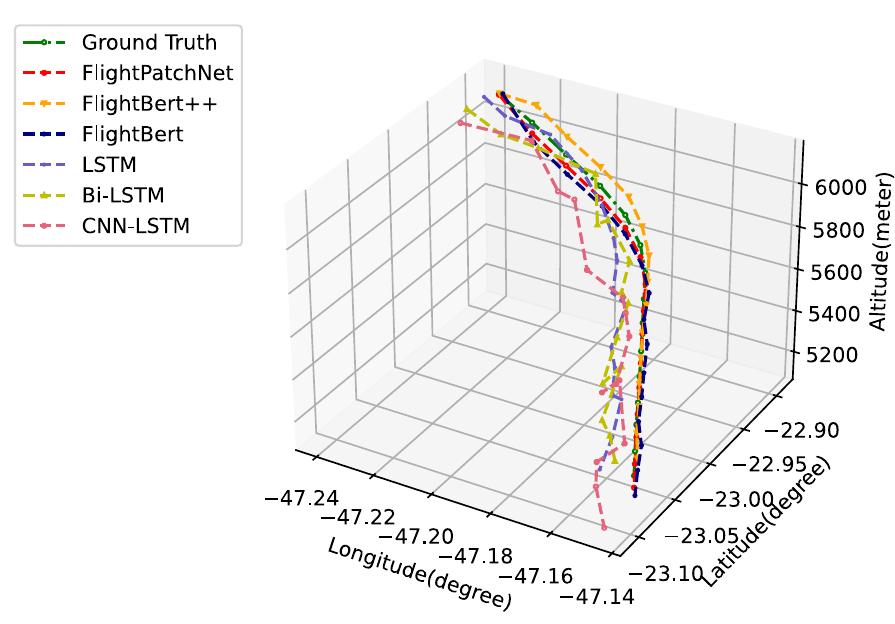}
    
    }\\
     \subfigure[]{
    
     \includegraphics[width=0.5\linewidth]{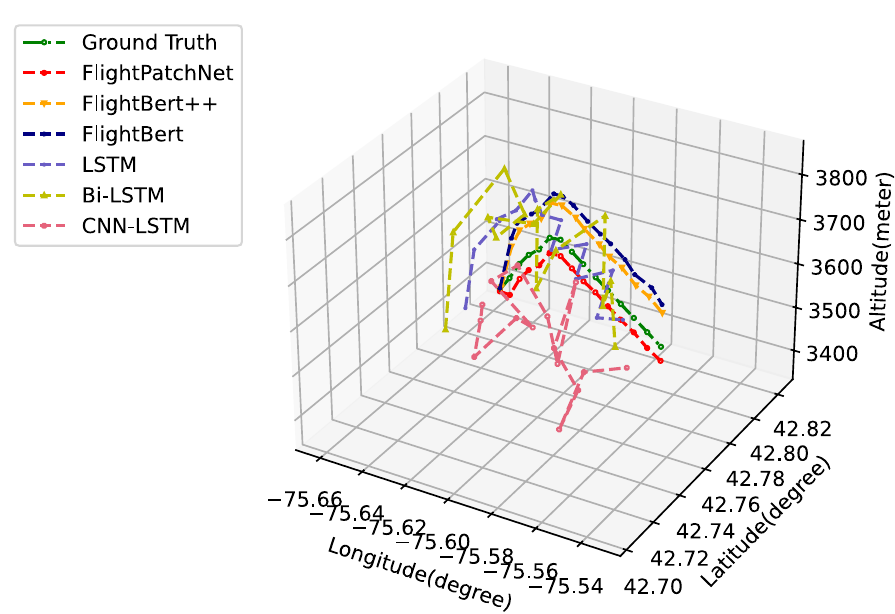}
    
    }
    \subfigure[]{
      \centering
     \includegraphics[width=0.5\linewidth]{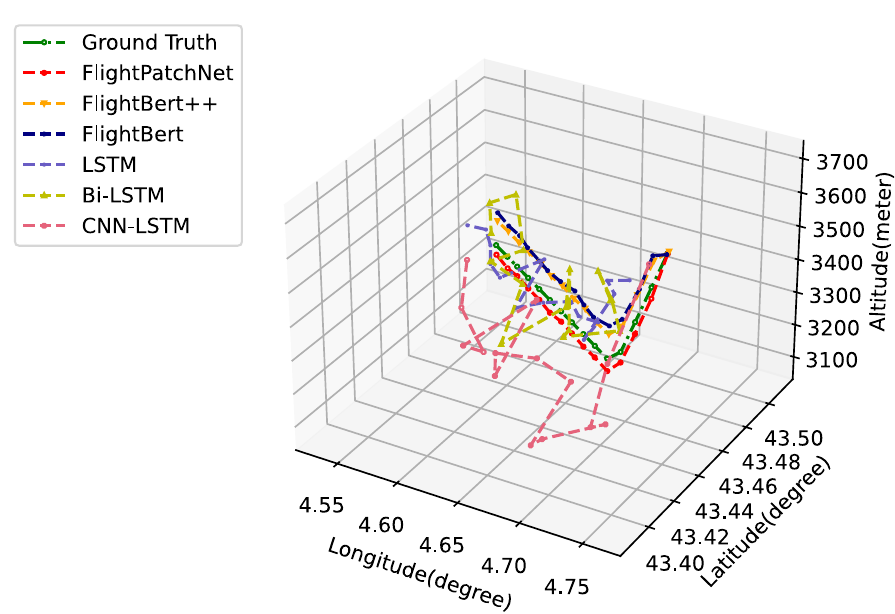}
    
    }  \\
     \subfigure[]{
      \centering
     \includegraphics[width=0.5\linewidth]{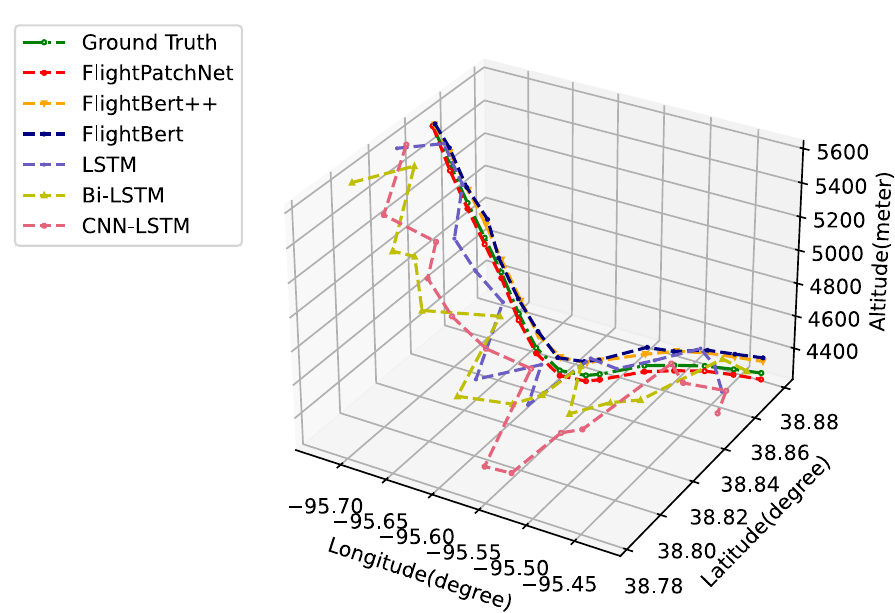}
    
    } 
    \subfigure[]{
      \centering
     \includegraphics[width=0.5\linewidth]{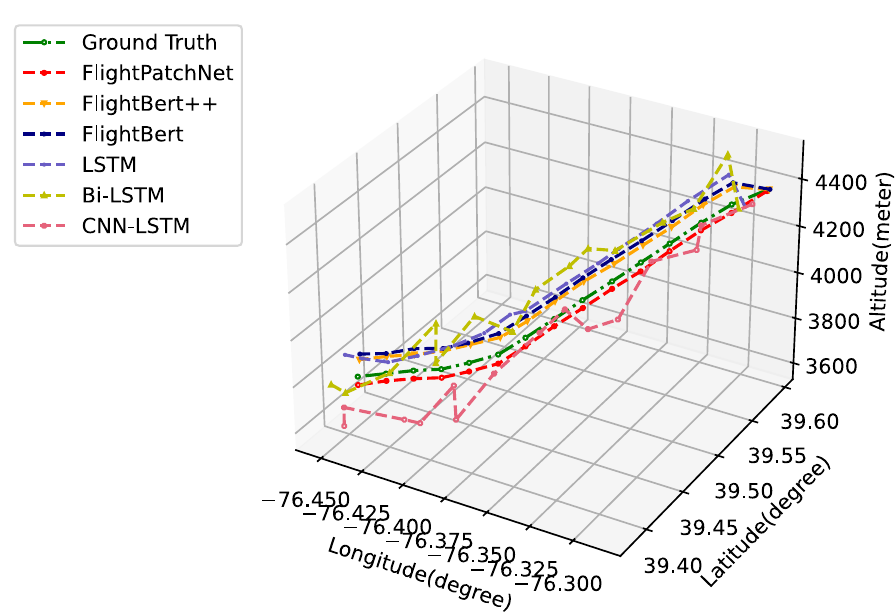}
    
    } 
    \captionsetup{font=small}
     \caption{Visualization of flight trajectory prediction results when the prediction horizon $T=15$ and look-back window size $L=60$.}
     \label{fig:traj_pred}
 \end{figure*}

 We visualize the flight trajectory prediction results of FlightPatchNet and all the baselines when the prediction horizon is 15. As shown in Figure~\ref{fig:traj_pred}, FlightPatchNet can provide stable and the most accurate predictions in longitude and latitude while it suffers from slight fluctuations in altitude. 

\end{document}